\pdfoutput=1

\documentclass[11pt,fleqn,leqno,a4paper]{article}

\usepackage[a4paper,left=17mm,right=17mm,
top=17mm,bottom=17mm]{geometry}

\usepackage[T1]{fontenc}  
\usepackage[utf8]{inputenc}

\usepackage{amssymb,amsmath,amsthm}
\usepackage{graphicx}

\newcommand{\kernn}[1]{\kern#1pt}
\newcommand{\raisee}[2]{\raisebox{#1}{\ensuremath{#2}}}

\newcommand{\mtiny}[1]{\mbox{\tiny\ensuremath{#1}}}

\newcommand{\mlargeeee}[1]{\mbox{\huge\ensuremath{#1}}}
\newcommand{\keraise}[3]{\kernn{#2}\raisee{#3pt}{\mtiny{#1}}}
\newcommand{\axstnn}[7]{\begin{array}{#7}\keraise{#1}{#2}{#3}\\\keraise{#4}{#5}{#6}\end{array}}
\newcommand{\bxstnn}[4]{\begin{array}[]{c}\mlargeeee{#1}\\[#4pt]\keraise{#2}{#3}{0}\end{array}}
\newcommand{\xconv}[5]{\kernn{-4}\axstnn{#4}{0}{0}{#3}{0}{0}{r}\kernn{-12}\bxstnn{\bb{C}}{#5}{8}{-18}
\kernn{-8}\axstnn{#2}{-2.5}{0}{#1}{-2.5}{0}{l}}
\newcommand{\xdense}[5]{\kernn{-6}\axstnn{#4}{0}{0}{#3}{0}{0}{r}\kernn{-14}
\bxstnn{\bb{F}}{#5}{16}{-17}\kernn{-17}\axstnn{#2}{5}{1}{#1}{2}{-1}{l}}

\newcommand{\xinp}[5]{\kernn{-4}\axstnn{#4}{0}{0}{#3}{0}{0}{r}\kernn{-15}
\bxstnn{\cl{I}}{#5}{12}{-17}\kernn{-14}\axstnn{#2}{5}{1}{#1}{2}{-1}{l}}
\newcommand{\xdrop}[5]{\kernn{-2}\axstnn{#4}{0}{0}{#3}{0}{0}{r}\kernn{-15}
\bxstnn{\bb{D}}{#5}{2}{-17}\kernn{-18}\axstnn{#2}{0}{0}{#1}{5}{-1}{l}}

\newtheorem{theorem}{Theorem}
\newcommand{\ovraa}[1]{\xrightarrow{#1}}
\newcommand{\ovlaa}[1]{\xleftarrow{#1}}

\newcommand{\bb}[1]{\mathbb{#1}}
\newcommand{\cl}[1]{\mathcal{#1}}
\newcommand{\tp}[1]{{#1}^{\intercal}}

\newcommand{\inv}[1]{\in\bb{R}^{#1}}

\def\ds{\displaystyle}
\def\ass{\leftarrow}
\def\od#1#2{\nabla_{#2}#1}

\def\cl#1{\mathcal{#1}}
\def\sp#1#2{\frac{\partial#1}{\partial#2}}

\def\eqd{\doteq}
\def\ra{\rightarrow}
\def\lra{\longrightarrow}

\def\xeq#1{\overset{#1}{=}}
\def\ov#1{\overline{#1}}

\def\diag#1{\mathtt{diag}\left[#1\right]}

\def\lra{\longrightarrow}

\def\sp#1#2{\frac{\partial{#1}}{\partial{#2}}}
\def\ov#1{\overline{#1}}
\def\ass{\leftarrow}
\def\ds{\displaystyle}
\def\eqd{\doteq}

\begin{document}
\title{Cross Entropy in Deep Learning of Classifiers Is Unnecessary\\- ISBE Error is All You Need}
\author{Władysław Skarbek\footnote{
Author is with the Faculty of Electronics and Information Technology, Warsaw University of Technology, email: wladyslaw.skarbek@pw.edu.pl
}}
\date{}

\maketitle

\begin{abstract}
In deep learning classifiers, the cost function usually takes the form of a combination of SoftMax and CrossEntropy functions. The SoftMax unit transforms the scores predicted by the model network into assessments of the degree (probabilities) of an object's membership to a given class. On the other hand, CrossEntropy measures the divergence of this prediction from the distribution of target scores. This work introduces the ISBE functionality, justifying the thesis about the redundancy of cross entropy computation in deep learning of classifiers. Not only can we omit the calculation of entropy, but also, during back-propagation, there is no need to direct the error to the normalization unit for its backward transformation. Instead, the error is sent directly to the model's network. Using examples of perceptron and convolutional networks as classifiers of images from the MNIST collection, it is observed for ISBE that results are not degraded with SoftMax only, but also with other activation functions such as Sigmoid, Tanh, or their hard variants HardSigmoid and HardTanh.
Moreover, up to three percent of time is saved within the total time of forward and backward stages. The article is addressed mainly to programmers and students interested in deep model learning. For example, it illustrates in code snippets possible ways to implement ISBE units, but also formally proves that the softmax trick only applies to the class of softmax functions with relocations.
\end{abstract}

\section{Introduction}

A deep model is a kind of mental shortcut (\cite{deep model}), broadly understood as a model created in deep learning of a certain artificial neural network $\cl{N}$, designed for a given application. What, then, is an artificial neural network (\cite{ann}), its deep learning (\cite{deep learning,deep learning RG}), and what applications (\cite{neural applications}) are we interested in?

From a programmer's perspective, an artificial neural network is a type of data processing algorithm (\cite{neural data processing}), in which subsequent steps are carried out by configurable computational units, and the order of processing steps is determined by a directed graph of connections without loops.

At the training stage, each group of input data $X$, i.e., each group of training examples, first undergoes the {\it inference (forward)} phase on the current model, i.e., processing through the network $\cl{N}$ at its current parameters $W$. As a result, network outputs $Y\ass\cl{F}_{\cl{N}}(X;W)$ are produced (\cite{inference}). 
\begin{equation*}
\begin{array}{l}
\ds\overbrace{\ovraa{X;\,W}\boxed{\cl{F}_{\cl{N}}}\ovraa{Y\ass\cl{F}_{\cl{N}}(X;W)}}^{\text{INFERENCE}} \quad\equiv\quad
\overbrace{
\boxed{
\quad\cdots\quad
\ovraa{X_u;\,W_u}
\boxed{\cl{F}_{\cl{U}}}
\ovraa{Y_u\ass\cl{F}_{\cl{U}}(X_u;W_u)}\quad\cdots\quad
}}^{\text{INFERENCE}}
\end{array}
\end{equation*}

After the inference phase comes the model update phase, where the current model is modified (improved) according to the selected optimization procedure (\cite{optim adam}).
{\it The model update phase} begins with calculating the loss (cost) value $Z\ass\cl{L}(Y,Y^{\circ})$ defined by the chosen loss function $\cl{L}$ as well as the inference outcome $Y$ and the target result $Y^{\circ}$.

\begin{equation*}
\begin{array}{l}
\overbrace{
\ds\ovraa{X;\,W}\boxed{\cl{F}_{\cl{N}}}\ovraa{Y\ass\cl{F}_{\cl{N}}(X;W)}}^{\text{INFERENCE}}\quad\cdots\quad
\overbrace{\ovraa{Y, Y^{\circ}}\boxed{\cl{L}}\ovraa{Z\ass\cl{L}(Y,Y^{\circ})}}^{\text{model update - start}}
\end{array}
\end{equation*}
The loss $Z$ depends (indirectly through $Y$) on all parameters $W$, and what conditions the next step of the update phase is the determination of sensitivity $\ov{W}$ of the loss function $\cl{L}$ to their changes. The mathematical model of sensitivity is the gradient $\ov{W}\doteq\sp{\cl{L}}{W}$. Knowing this gradient, the optimizer will make the actual modification of $W$ in a direction that also takes into account the values of gradients obtained for previous training batches.

Calculating the gradient with respect to parameters actually assigned to different computational units required the development of an efficient algorithm for its propagation in the opposite direction to inference (\cite{backward,backward NN}).

Just as in the inference phase, each unit $\cl{U}$ has its formula $Y_u\ass\cl{F}_{\cl{U}}(X_u,W_u)$ for processing data from input $X_u$ to output $Y_u$ with parameters $W_u$, so in the {\it backward gradient propagation phase}, it must have a formula $\ov{X}_u,\ov{W}_u\ass\ov{\cl{F}}_{\cl{U}}(\ov{Y}_u)$ for transforming the gradients assigned to its outputs $\ov{Y}_u$ into gradients assigned to its inputs $\ov{X}_u$ and its parameters $\ov{W}_u$. 
\begin{equation*}
\begin{array}{c}
\overbrace{
\ovlaa{\ov{X};\,\ov{W}\ass\ov{\cl{F}}_{\cl{N}}(\ov{Y};X,Y,W)}\boxed{\ov{\cl{F}}_{\cl{N}}}\ovlaa{\ov{Y}}}^{\text{BACKPROPAGATION}}\quad\cdots\quad
\overbrace{\ovlaa{\ov{Y}\ass\ov{\cl{L}}(\ov{Z};Y,Z)}\boxed{\ov{\cl{L}}}\ovlaa{\mathbf{1}=\sp{Z}{Z}}}^{\text{loss function gradient\,}\sp{L}{Y}}\\[10pt]
\overbrace{
\boxed{
\quad\cdots\quad
\ovlaa{\ov{X}_u;\,\ov{W}_u\ass\ov{\cl{F}}_{\cl{U}}(\ov{Y}_u;X_u,Y_u,W_u)}\boxed{\ov{\cl{F}}_{\cl{U}}}\ovlaa{\ov{Y}_u}\quad\cdots\quad
}}^{\text{GRADIENT BACKPROPAGATION}}
\end{array}
\end{equation*}
Based on such local rules of gradient backpropagation and the created computation graph, the {\it backpropagation} algorithm can determine the gradients of the cost function with respect to each parameter in the network. The computation graph is created during the inference phase and is essentially a stack of links between the arguments and results of calculations performed in successive units (\cite{backward NN,pytorch ad}).

Deep learning is precisely a concert of these inference and update phases in the form of gradient propagation, calculated for randomly created groups of training examples. These phases, intertwined, operate on multidimensional deep tensors (arrays) of data, processed with respect to network inputs and on deep tensors of gradient data, processed with respect to losses, determined for the output data of the trained network.

Here, by a deep tensor, we mean a multidimensional data array that has many feature maps, i.e., its size along the feature axis is relatively large, e.g., 500, which means 500 scalar feature maps. We then say that at this point in the network, our data has a {\it deep representation} in a 500-dimensional space.

As for the applications we are interested in this work, the answer is those that have at least one requirement for classification (\cite{class task}). An example could be crop detection from satellite images (\cite{class crop}), building segmentation in aerial photos \cite{air segment}, but also text translation (\cite{NLP}). Classification is also related to voice command recognition (\cite{voice commands}), speaker recognition (\cite{speaker recognition}), segmentation of the audio track according to speakers (\cite{speaker diarization}), recognition of speaker emotions with visual support (\cite{visual speech}), but also classification of objects of interest along with their localization in the image (\cite{detection}).

It may be risky to say that after 2015, in all the aforementioned deep learning classifiers, the cost function takes the form of a composition of the $SoftMax$ function (\cite{softmax}) and the $CrossEntropy$ function, i.e., cross-entropy (\cite{cross entropy}). The {\tt SoftMax} unit normalizes the scores predicted by the classifier model for the input object, into {\it softmax} scores that sum up to one, which can be treated as an estimation of the conditional probability distribution of classes. Meanwhile, cross-entropy measures the divergence (i.e., divergence) of this estimation from the target probability distribution (class scores). In practice, the target score may be taken from a training set prepared manually by a so-called teacher (\cite{supervised learning}) or may be calculated automatically by another model component, e.g., in the knowledge distillation technique (\cite{knowledge distillation}).

For $K$ classes and $n_b$ training examples, the $SoftMax$ function is defined for the raw score matrix $X\in\bb{R}^{n_b\times K}$ as:
\begin{equation*}
\left[Y \ass SoftMax(X)\right]\quad\lra\quad 
 \left[Y_{bi}\ass
\frac{e^{X_{bi}}}{\ds\sum_{j\in[K]}e^{X_{bj}}},\, b\in[n_b],\, i\in[K]\right]\,,
\end{equation*}
where the notation $[K]$ denotes any $K$-element set of indices - in this case, they are class labels.

The $CrossEntropy$ function on the matrix $Y,Y^{\circ}\in\bb{R}^{n_b\times K}$ is defined by the formula:
$$
\left[Z \ass CrossEntropy(Y,Y^{\circ})\right] \lra
\left[Z_b\ass\ds -\sum_{j\in[K]}Y^{\circ}_{bj}\ln Y_{bj},\, b\in[n_b],\, z\in\bb{R}^{n_b}\right]
$$

\begin{equation}
\label{SOCE:sep}
\overbrace{
\begin{array}{l}
\overbrace{
\ovraa{\text{classified object}}
\boxed{\cl{F}_{\cl{N}}}
\ovraa{\text{raw scores\,}X}}^{
\text{SCORES INFERENCE}}\\[5pt]
\overbrace{
\ovraa{\text{raw scores\,}X}
\boxed{\text{\tt SoftMax}}
\ovraa{\text{soft scores\,}Y,\,Y^{\circ}}
\boxed{\text{\tt CrossEntropy}}
\ovraa{\text{losses\,}Z}}^{\text{LOSS ESTIMATION\,}\cl{L}}
\end{array}
}^{\text{Classifier loss function: Separated Implementation}}
\end{equation}

When classifiers began using a separated implementation of the combination of the normalization unit {\tt SoftMax} and the unit {\tt CrossEntropy}, it quickly became evident in practice that its implementation had problems with scores close to zero, both in the inference phase and in the backward propagation of its gradient. Only the integration of {\tt CrossEntropy} with normalization {\tt SoftMax} eliminated these inconveniences. The integrated approach has the following form:
\begin{equation*}
\overbrace{
\begin{array}{l}
\overbrace{
\ovraa{\text{classified object}}
\boxed{\cl{F}_{\cl{N}}}
\ovraa{\text{raw scores\,}X}}^{\text{INFERENCE}}
\\[7pt]
\overbrace{
\ovraa{\text{raw scores\,}X,\text{soft scores\,}Y^{\circ}}
\boxed{\text{\tt CrossEntropy\ $\circ$\ SoftMax}}
\ovraa{\text{losses\,}Z}}^{\text{LOSS ESTIMATION\,}\cl{L}}
\end{array}
}^{\text{Classifier loss function - Integrated Implementation}}
\end{equation*}

The integrated functionality of these two features has the following redundant mathematical notation:
\begin{equation*}\label{SOCE: int}
\begin{array}{l}
Z \ass \left[CrossEntropy\, \circ\, SoftMax\right](X,Y^{\circ}) \lra\\[5pt]
Z_b\ass\ds -\sum_{j\in[K]}
Y^{\circ}_{bj}\ln \frac{e^{X_{bj}}}{\ds\sum_{i\in[K]}
e^{X_{bi}}},\, b\in[n_b]
\end{array}
\end{equation*}
This redundancy in notation was helpful in deriving the equation for the gradient backpropagation for the integrated loss function 
$CrossEntropy\,\circ\,SoftMax$ (\cite{historia strick}).

The structure of this paper is as follows:
\begin{enumerate}
\item
In the second section titled {\it ISBE Functionality}, the conditions that a normalization unit must meet for its combination with a cross-entropy unit to have a gradient at the input equal to the difference in soft scores: $\ov{X} = Y-Y^{\circ}$ are analyzed. Then the definition of ISBE functionality is introduced, which in the inference phase {\tt(I)} normalizes the raw score to a soft score {\tt(S)}, and in the backward propagation phase {\tt(B)} returns an error {\tt(E)}, equal to the difference in soft scores. It is also justified why in the case of the $SoftMax$ normalization function, the {\tt ISBE} unit has, from the perspective of the learning process, the functionality of the integrated unit {\tt CrossEntropy\ $\circ$\ SoftMax}.
\item In the third section, using the example of the problem of recognizing handwritten digits and the standard {\tt MNIST(60K)} image collection (\cite{mnist}), numerous experiments show that in addition to the obvious savings in computational resources, in the case of five activations serving as normalization functions, the classifier's effectiveness is not lower than that of the combination of the normalization unit {\tt Softmax} and the unit {\tt Cross Entropy}. This {\tt ISBE} property was verified for the activation units {\tt Softmax, Sigmoid, Hardsigmoid,} and {\tt Tanh} and {\tt Hardtanh}.
\item The last fourth section contains conclusions.
\item Appendix A, titled {\it Cross-Entropy and Softmax Trick Properties}, contains the formulation and proof of the theorem on the properties of the {\it softmax trick}.
\end{enumerate}

\section{ISBE Functionality}

The ISBE functionality is a proposed simplification of the cost function, combining the SoftMax normalization function with the cross-entropy function, hereafter abbreviated as CE$_{all}$. Its role is to {\it punish} those calculated probability distributions that significantly differ from the distributions of scores proposed by the {\it teacher}.

To understand this idea, let's extend the inference diagram for CE$_{all}$ with the backward propagation part for the gradient. We consider this diagram in its separated version, omitting earlier descriptions for the diagram \eqref{SOCE:sep}:
\begin{equation}\label{diag:SOCE}
\begin{array}{l}
\overbrace{
\ovraa{X}
\boxed{\text{\tt SoftMax}}
\ovraa{Y,\,Y^{\circ}}
\boxed{\text{\tt CrossEntropy}}
\ovraa{Z}}^{\text{LOSS INFERENCE\,}\cl{L}}\\[10pt]
\overbrace{
\ovlaa{\ov{X}\ovlaa{Theorem\,\ref{th:strick}}Y-Y^{\circ}}
\boxed{\ov{\text{\tt SoftMax}}}
\ovlaa{\ov{Y};\,Y,\,Y^{\circ}}
\boxed{\ov{\text{\tt CrossEntropy}}}
\ovlaa{\ov{Z}=\mathbf{1}}}^{\text{BACKPROPAGATION}}
\end{array}
\end{equation} 
  
The meaning of variables $X,Y,Y^{\circ},Z$ and $\ov{Z},\ov{Y},\ov{X}$ appearing in the above diagram \eqref{diag:SOCE}:\\
$
\begin{array}{ll}
 X & \text{raw score at the input of the normalization function preceding cross-entropy CE}, X\in\bb{R}^K,\\
 Y & \text{normalization result, so-called soft score\,,} Y\in(0,1)^K,\\
 Y^{\circ} & \text{target soft score, assigned to the classified example\,,}\\
 Z & \text{output of cross-entropy CE\,,} Z\in\bb{R},\\
 \ov{Z} & \text{formal gradient at the input of the backward propagation algorithm,\,} \ov{Z}=1,\\
 \ov{Y} & \text{gradient of cross-entropy CE with respect to Y:\,} \ov{Y} = \sp{Z}{Y}=-\frac{Y^{\circ}}{Y},\\[3pt]
 \ov{X} & \text{gradient of cross-entropy CE with respect to X:\,} 
 \ov{X}\ovlaa{Theorem\, \ref{th:strick}}(Y-Y^{\circ})\,.
\end{array}
$

The key formula here is $\ov{X}\ass(Y-Y^{\circ})$. Its validity comes from the mentioned theorem and the formula \eqref{eq:strick} associated with the {\it softmax trick} property.

The equation \eqref{eq:jacobian} on the form of the Jacobian of the normalization unit is both a sufficient and necessary condition for its combination with the cross-entropy unit to ensure the equality \eqref{eq:strick}. Moreover, this condition implies that an activation function with a Jacobian of the {\it softmax} type is a {\tt SoftMax} function with optional relocation.

Theorem \ref{th:strick} leads us to a seemingly pessimistic conclusion: it is not possible to seek further improvements by changing the activation and at the same time expect the {\it softmax trick} property to hold. Thus, the question arises: what will happen if, along with changing the activation unit, we change the cross-entropy unit to another, or even omit it entirely?

In the {\it ISBE} approach, the aforementioned simplification of the CE$_{all}$ cost function involves precisely omitting the cross-entropy operation in the inference stage and practically omitting all backward operations for this cost function. So what remains? The answer is also an opportunity to decode the acronym {\it ISBE} again:
\begin{enumerate}
\item In the inference phase {\tt(I)}, we normalize the raw score $X$ to $Y=SoftMax(X)$, characterized as a soft score {\tt(S)}.
\item In the backward propagation phase {\tt(B)}, we return an error {\tt(E)} equal to the difference between the calculated soft score and the target score, i.e., 
$\ov{X}\doteq Y-Y^{\circ}$.
\end{enumerate}

Why can we do this and still consider that in the case of the {\tt SoftMax} activation function, the value of the gradient transmitted to the network is identical: $\ov{X}_{CE_{all}} = \ov{X}_{ISBE}\eqd Y-Y^{\circ}$?  

The answer comes directly from the property $\ov{X}_{CE_{all}} = Y-Y^{\circ}$, formulated in equation \eqref{eq:strick}, which was defined in the theorem \ref{th:strick} as the {\it softmax trick} property.

We thus have on the left the following diagram of data and gradient backpropagation through such a unit. On the right we have its generalization to a  {\tt ScoreNormalization} unit instead of {\tt SoftMax} unit.
\begin{equation*}\label{isbe:a}
\left.
\begin{array}{l}
\overbrace{
\ovraa{X}
\boxed{\text{\tt SoftMax}}
\ovraa{Y,\,Y^{\circ}}}^{\text{ISBE INFERENCE}}\\[7pt]
\overbrace{
\ovlaa{\ov{X}\ass Y-Y^{\circ}}
\boxed{{\text{\tt Subtract}}}
\ovlaa{Y,\,Y^{\circ}}}
^{\text{$\ov{ISBE}$\, BACKPROPAGATION}}
\end{array}
\right\}
\ovraa{\text{generalize}}
\left\{
\begin{array}{l}
\overbrace{
\ovraa{X}
\boxed{\text{\tt Score\,Normalization}}
\ovraa{Y,\,Y^{\circ}}}^{\text{ISBE INFERENCE}}\\[7pt]
\overbrace{
\ovlaa{\ov{X}\ass Y-Y^{\circ}}
\boxed{{\text{\tt Subtract}}}
\ovlaa{Y,\,Y^{\circ}}}
^{\text{$\ov{ISBE}$\, BACKPROPAGATION}}
\end{array}
\right.
\end{equation*}

Which activation functions should we reach for in order to test them in the ISBE technique?
\begin{enumerate}
\item The SoftMax activation function should be the first candidate for comparison, as it theoretically guarantees behavior comparable to the system containing cross-entropy.
\item Activations should be monotonic, so that the largest value of the raw score remains the largest score in the soft score sequence.
\item Soft scores should be within a limited range, e.g., $[0,1]$ as in the case of SoftMax and Sigmoid, or $[-1,+1]$ as for Tanh.
\item The activation function should not map two close scores to distant scores. For example, normalizing a vector of scores by projecting onto a unit sphere in the $p$-th Minkowski norm meets all above conditions, however, it is not stable around zero. Normalization $\frac{x}{\|x\|_p}$ maps, for example, two points $\epsilon,-\epsilon$ distant by $2\cdot\|\epsilon\|_p$ to points distant exactly by $2$, thus changing their distance $\frac{1}{\|\epsilon\|_p}$ times, e.g., a million times, when $\|\epsilon\|_p=10^{-6}$. This operation is known in {\tt Pytorch} library as {\tt normalize} function.
\end{enumerate}

The experiments conducted confirm the validity of the above recommendations. The {\tt Pytorch} library functions {\tt softmax, sigmoid, tanh, hardsigmoid, hardtanh} meet the above three conditions and provide effective classification at a level of effectiveness higher than $99.5\%$, comparable to {\tt CrossEntropy $\circ$ SoftMax}. In contrast, the function {\tt normalize} gave results over $10\%$ worse - on the same {\tt MNIST(60K)} collection and with the same architectures.

What connects these {\it good normalization} functions $F:\bb{R}^K\ra\bb{R}^K$, of which two are not even fully differentiable? Certainly, it is the Lipschitz condition occurring in a certain neighborhood of zero (\cite{Lipshitz}):
$$
 x\in\bb{R}^K,\ \|x\|_p\leq\epsilon \lra \|F(x)\|_p\leq c\|x\|_p\,,\quad\text{where $c$ is a certain constant\,.}
$$
Note that the Lipschitz condition meets the expectations of the fourth requirement on the above list of recommendations for {\tt ISBE}. Moreover, we do not expect here that the constant $c$ be less than one, i.e., that the function $F$ has a narrowing character.

We need also a recommendation for {\it teachers} preparing class labels, which we represent as vectors blurred around the base vectors of axes $I_K=[e_1,\dots,e_K],\ e_i[j]\eqd\delta_{ij}$:
\begin{enumerate}
\item example blurring value $\mu$, e.g., $\mu=10^{-6}$:
$$
\tilde{e}_i[j] \ass (1-\mu)\delta_{ij} + \frac{\mu}{K-1}(1-\delta_{ij})
$$
\item when the range of activation values is other than the interval $[0,1]$, we adjust the vector $\tilde{e}_i$ to the new range, e.g., for {\tt tanh} the range is the interval $(-1,+1)$ and then the adjustment has the form:
$$
\tilde{e}_i \ass 2\cdot\tilde{e}_i-1,\ i=1,\dots,K
$$
\end{enumerate}

Finally, let's take a look at the code for the main loop of the program implemented on the {\tt Pytorch} platform. 
\begin{enumerate}
\item This is what the code looks like when {\tt loss\_function} is chosen as {\tt nn.CrossEntropyLoss}:
\begin{verbatim}
for (labels,images) in tgen:
    outputs = net(images)
    loss = loss_function(outputs, labels)
    optimizer.zero_grad()
    loss.backward()
    optimizer.step()
\end{verbatim}
\item Now we introduce the {\tt ISBE} option for {\tt SoftMax} activation:
\begin{verbatim}
for (labels,images) in tgen:
    outputs = net(images)
    if no_cross_entropy:
        soft_error = softmax(outputs) - labels
        optimizer.zero_grad()
        outputs.backward(soft_error)
    else:
        loss = loss_function(outputs, labels)
        optimizer.zero_grad()
        loss.backward()
    optimizer.step()
\end{verbatim}
\item If we want to test more options, the loop code will extend a bit:
\begin{verbatim}
for (labels,images) in tgen:
    outputs = net(images)
    if no_cross_entropy:
        if soft_option=="softmax":
            soft_error = softmax(outputs) - labels
        if soft_option=="tanh":
            soft_error = tanh(outputs) - (2.*labels-1.)
        elif soft_option=="hardtanh":
            soft_error = hardtanh(outputs) - (2.*labels-1.)
        elif # ...
             # next options
        optimizer.zero_grad()
        outputs.backward(soft_error)
    else:
        loss = loss_function(outputs, labels)
        optimizer.zero_grad()
        loss.backward()
    optimizer.step()
\end{verbatim}
\item If we prefer to have a visually shorter loop, then by introducing the variable {\tt soft\_function} and extending the class {\tt DataProvider} with matching target labels for a given soft option, we finally get a compact form:
\begin{verbatim}
for (labels,images) in tgen:
    outputs = net(images)
    if no_cross_entropy:
        soft_error = soft_function(outputs) - labels
        optimizer.zero_grad()
        outputs.backward(soft_error)
    else:
        loss = loss_function(outputs, labels)
        optimizer.zero_grad()
        loss.backward()
    optimizer.step()
\end{verbatim}
\end{enumerate}

Of course, the above code snippets are only intended to illustrate how easy it is to add the functionality of {\tt ISBE} to an existing application.

\section{Experiments}

What do we want to learn from the planned experiments? We already know from theory that in the case of the {\tt SoftMax} activation, we cannot worsen the parameters of the classifier using cross-entropy, both in terms of success rate and learning time.

Therefore, we first want to verify whether theory aligns with practice, but also to check for which normalization functions the ISBE unit does not degrade the model's effectiveness compared to CE$_{all}$.

The learning time $t_{ISBE}$ should be shorter than $t_{CE}$, but to be independent of the specific implementation, we will compare the percentage of the backpropagation time in the total time of inference and backpropagation:
\begin{equation}\label{eq:tau}
\tau \eqd \frac{\text{backpropagation time}}{
\text{inference time} + \text{backpropagation time}}\cdot 100\%
\end{equation}

We evaluate the efficiency of the ISBE idea on the standard {\tt MNIST(60K)} image collection and the problem of their classification.

From many quality metrics, we choose the success rate (also called accuracy), defined as the percentage of correctly classified elements from the test collection {\tt MNIST(10K)}
\begin{equation}\label{eq:alpha}
\alpha  = \frac{\text{number of correct classifications}}{\text{size of the test collection}}\cdot 100\%
\end{equation}

We want to know how this value changes when we choose different architectures, different activations in the ISBE technique, but also different options for aggregating cross-entropy over the elements of the training batch.
Thus, we have the following degrees of freedom in our experiments:

\begin{enumerate}
\item Two architecture options  
\begin{itemize}
\item Architecture $\cl{N}_0$ consists of two convolutions\quad $\xconv{}{}{}{}{}$ and two linear units\quad $\xdense{}{}{}{}{}$, of which the last one is a projection from the space of deep feature vectors of dimension $512$ to the space of raw scores for each of the $K=10$ classes:  
$$
\ovraa{\text{\small image}}
\xinp{\kernn{-2}28_{yx}}{1}{}{}{}
\xconv{3^k2^s}{32}{}{}{}
\xconv{3^k2^s}{64}{}{}{}
\xdrop{20}{}{}{}{}
\xdense{}{512}{}{}{}
\xdense{}{10}{}{}{}
\ovraa{\text{\small class\ scores}}
$$
as by STNN notation (\cite{stnn}), for instance\\
 $\left\{
\begin{array}{cl}
\xconv{3^k2^s}{32}{}{}{} & \text{means 32 convolutions with 3x3 masks, sampled with a stride of 2},\\
\xdrop{20}{}{}{}{} & \text{DropOut - a unit zeroing 20\% of tensor elements},\\
\xdense{}{512}{}{}{} & \text{a linear unit with a matrix $A\in\bb{R}^{?\times512}$},\\ & \text{here $?=64$ - it is derived from the shape of}\\ &\text{the tensor produced by the previous unit\,.}
\end{array}
\right.$ 
\item  Architecture $\cl{N}_1$ consists of two blocks, each with 3 convolutions - it is a purely convolutional network, except for the final projection:
$$
\begin{array}{l}
\ovraa{\text{\small image}}
\xinp{\kernn{-2}28_{yx}}{1}{}{}{}
\ \xconv{3^k}{32}{}{}{}
\xconv{3^k}{32}{}{}{}
\xconv{5^k2^s}{32}{p}{}{}
\ \xdrop{40}{}{}{}{}
\xconv{3^k}{64}{}{}{}
\xconv{3^k}{64}{}{}{}
\xconv{5^k2^s}{64}{p}{}{}
\ \xdrop{40}{}{}{}{}
\xconv{4^k}{128}{}{}{}
\xdense{}{10}{}{}{}
\ovraa{\text{\small class\ scores}}\\
\end{array}
$$
Note that the last convolution in each block has a {\tt p} requirement for padding, i.e., filling the domain of the image with additional lines and rows so that the image resolution does not change.
\end{itemize}
\item Three options for reducing the vector of losses in the {\tt CrossEntropyLoss} unit:
{\tt none, mean, sum}. 
\item Five options for activation functions used in the ISBE technique:
\begin{itemize}
\item {\tt SoftMax}: $\ds y_i \ass \frac{e^{x_i}}{\ds\sum_{j\in[K]} e^{x_j}},\quad i\in[K],$
\item {\tt Tanh}:  $\ds y_i \ass \frac{e^{x_i}-e^{-x_i}}{e^{x_i}+e^{-x_i}},\quad i\in[K],$
\item {\tt HardTanh}: 
$\ds
y_i\ass \left\{
\begin{array}{cl}
-1 & \text{if } x_i\leq -1\\
x_i  & \text{if } -1 < x_i < +1\\
+1 & \text{if } +1\leq x_i
\end{array}\right\},\quad
i\in[K],
$
\item {\tt Sigmoid}: $\ds y_i \ass \frac{1}{1+e^{-x_i}},\ i\in[K],$
\item {\tt HardSigmoid}: 
$\ds y_i \ass
\left\{
\begin{array}{cl}
0 & \text{gdy } x_i\leq-2\\
\frac{x_i+2}{4}  & \text{gdy } -2 < x_i < +2\\
+1 & \text{gdy } +2\leq x_i
\end{array}\right\} 
= \frac{HardTanh(x_i/2)+1}{2},\quad i\in[K].$
\end{itemize}
\end{enumerate}

The results of the experiments, on the one hand, confirm our assumption that the conceptual Occam's razor, i.e., the omission of the cross-entropy unit, results in time savings $\tau$, and on the other hand, the results are surprisingly positive with an improvement in the metric of success rate $\alpha$ in the case of hard activation functions $HardTanh$ and $HardSigmoid$. It was observed that only the option of reduction by {\tt none} behaves exactly according to theory, i.e., the success rate is identical to the model using $SoftMax$ normalization. Options {\tt mean} and {\tt sum} for the model with {\it entropy} are slightly better than the model with {\it softmax}. 

The consistency of models in this case means that the number of images incorrectly classified out of 10 thousand is the same. In the experiments, it was not checked whether it concerns the same images. A slight improvement, in this case, meant that there were less than a few or a dozen errors, and the efficiency of the model above $99.6\%$ meant at most $40$ errors per $10$ thousand of test images.

\subsection{Comparison of time complexity}

We compare time complexity according to the metric given by the formula \eqref{eq:tau}.

\begin{table}[ht]
\caption{Comparison of the metric $\tau$, i.e., the percentage share of backpropagation time in the total time with inference. The share $\tau_{CE}$ of cross-entropy with three types of reduction is compared with five functions of soft normalization. The analysis was performed for architectures $\cl{N}_0$ and $\cl{N}_1$.}\label{tab:time-all}
\centerline{
\begin{tabular}{c||c|c|c||c|c|c|c|c}
net & mean & none & sum & hsigmoid & htanh & sigmoid & softmax & tanh
\\\hline
$\cl{N}_0$ & 60.61\% & 59.56\% & 59.98\% & 58.31\% & 58.21\% & 57.38\% & 57.45\% & 59.07\%\\\hline
$\cl{N}_1$ & 54.89\%& 53.92\%& 53.98\%& 52.68\%& 52.33\%& 51.75\%& 51.95\%& 52.11\%
\\\hline
$\cl{N}_1^r$ 
& 54.45\%
& 53.92\%
& 54.00\%
& 52.78\%
& 52.30\%
& 51.67\%
& 51.73\%
& 52.11\%
\end{tabular}
}
\end{table}

\begin{table}[ht]
\caption{Metric $\Delta\tau\eqd\tau_{ISBE}- \tau_{CE}$, i.e., the decrease in the percentage share of backpropagation time in the total time with inference. The analysis was performed for architectures $\cl{N}_0$ and $\cl{N}_1$.}\label{tab:time-delta}
\centerline{
\begin{tabular}{c||c||c|c|c|c|c}
net & CE loss & hsigmoid & htanh & sigmoid & softmax & tanh
\\\hline
$\cl{N}_0$ & mean & -2.30\%& -2.40\%& -3.23\%& -3.16\%& -1.54\%\\
$\cl{N}_0$ & none & -1.25\%& -1.35\%& -2.18\%& -2.11\%& -0.50\%\\
$\cl{N}_0$ & sum  & -1.67\%& -1.77\%& -2.60\%& -2.53\%& -0.92\%\\\hline
$\cl{N}_1$ & mean & -2.21\%& -2.56\%& -3.14\%& -2.94\%& -2.79\%\\
$\cl{N}_1$ & none & -1.24\%& -1.59\%& -2.17\%& -1.97\%& -1.82\%\\
$\cl{N}_1$ & sum  & -1.30\%& -1.65\%& -2.23\%& -2.03\%& -1.87\%
\end{tabular}
}
\end{table}
In the context of time, the table \ref{tab:time-all} clearly shows that the total time share of backpropagation, obviously depending on the complexity of the architecture, affects the time savings of the ISBE technique compared to {\tt CrossEntropyLoss} - table 
\ref{tab:time-delta}. The absence of pluses in this table, i.e., the fact that all solutions based on {\tt ISBE} are relatively faster in the learning phase, is an undeniable fact.

The greatest decrease in the share of backpropagation, over $3\%$, occurs for the $Sigmoid$ and $SoftMax$ activations. The smallest decrease in architecture $\cl{N}_0$ is noted for the soft (soft) normalization function $Tanh$ and its hard version $HardTanh$. This decrease refers to cross-entropy without reduction, which is an aggregation of losses calculated for all training examples in a given group, into one numerical value.

Inspired by the theorem  \ref{th:strick} , which states that the relocation of the $SoftMax$ function preserves the {\it softmax trick} property, we also add data to the table \ref{tab:time-all} for the network $\cl{N}_1^r$. This network differs from the $\cl{N}_1$ network only in that the normalization unit has a trained relocation parameter. In practice, we accomplish training with relocation for normalization by training with the relocation of the linear unit immediately preceding it. This is done by setting its parameter: {\tt bias=True}.

As we can see, the general conclusion about the advantage of the ISBE technique in terms of time reducing for the model with the relocation of the normalization function, is the same.

\subsection{Comparison of classifier accuracy}
 
Comparison of classifier accuracy and differences in this metric are contained in tables \ref{tab:acc-all} and \ref{tab:acc-delta}. 
The accuracy is computed according the formula \eqref{eq:alpha}.

The number of pluses on the side of {\tt ISBE} clearly exceeds the number of minuses. The justification for this phenomenon requires separate research. Some light will be shed on this aspect by the analysis of learning curves - the variance in the final phase of learning is clearly lower. The learning process is more stable.

\begin{table}[ht]
\caption{In the table, the success rate of three classifiers based on cross-entropy with different aggregation options is compared with the success rate determined for five options of soft score normalization functions. The analysis was performed for architectures $\cl{N}_0$ and $\cl{N}_1$.}\label{tab:acc-all}
\centerline{
\begin{tabular}{c||c|c|c||c|c|c|c|c}
net & mean & none & sum & hsigmoid & htanh & sigmoid & softmax & tanh
\\\hline
$\cl{N}_0$ & 99.45\%& 99.41\%& 99.47\%& 99.50\%& 99.50\%& 99.56\%& 99.41\%& 99.45\%\\\hline
$\cl{N}_1$ & 99.61\%& 99.58\%& 99.59\%& 99.64\%& 99.66\%& 99.64\%& 99.62\%& 99.63\%
\\\hline
$\cl{N}_1^r$ 
& 99.55\%
& 99.64\%
& 99.64\%
& 99.61\%
& 99.66\%
& 99.69\%
& 99.63\%
& 99.57\%
\end{tabular}
}
\end{table}

\begin{table}[ht]
\caption{Change in success rate between models with cross-entropy and models with soft score normalization function. The analysis was performed for architectures $\cl{N}_0$ and $\cl{N}_1$.}\label{tab:acc-delta}
\centerline{
\begin{tabular}{c||c||c|c|c|c|c}
net & CE loss & hsigmoid & htanh & sigmoid & softmax & tanh
\\\hline
$\cl{N}_0$ & mean & 0.05\%& 0.05\%& 0.11\%& -0.04\%& 0.00\%\\
$\cl{N}_0$ & none & 0.09\%& 0.09\%& 0.15\%& 0.00\%& 0.00\%\\
$\cl{N}_0$ & sum  & 0.13\%& 0.03\%& 0.09\%& -0.06\%& -0.02\%\\\hline
$\cl{N}_1$ & mean & 0.03\%& 0.05\%& 0.03\%& 0.01\%& 0.02\%\\
$\cl{N}_1$ & none & 0.06\%& 0.08\%& 0.06\%& 0.04\%& 0.05\%\\
$\cl{N}_1$ & sum  & 0.05\%& 0.07\%& 0.05\%& 0.03\%& 0.04\%
\end{tabular}
}
\end{table}

In the table \ref{tab:acc-delta}, we observe that, with the exception of the function $SoftMax$, which on several images of digits performed worse than the model with cross-entropy, the soft activations have an efficiency slightly or significantly better.
However, we are talking about levels of tenths or hundredths of a percent here. The largest difference noted for the option {\tt softmax} was $15$ hundredths of a percent, meaning $15$ more images correctly classified. Such differences are within the margin of statistical error.

The use of relocation for the normalization functionof does not provide a clear conclusion - for some models there is a slight improvement, for others a slight deterioration. It is true that the {\tt ISBE} unit with {\tt sigmoid} activation achieved the best efficiency of $99.69\%$, but this is only a matter of a few images.

Within the limits of statistical error, we can say that the {\tt ISBE} technique gives the same results in recognizing {\tt MNIST} classes. Its advantages are:
\begin{itemize}
\item  of decrease time in the total time,
\item simplification of architecture, and therefore plaing the philosophical role of {\it Occam's razor}.
\end{itemize}

\subsection{Visual analysis}

Further analysis of the results will be based on the visual comparison of learning curves. 

\begin{figure}[ht]
\centerline{
\begin{tabular}{ccc}
\includegraphics[height=120mm]{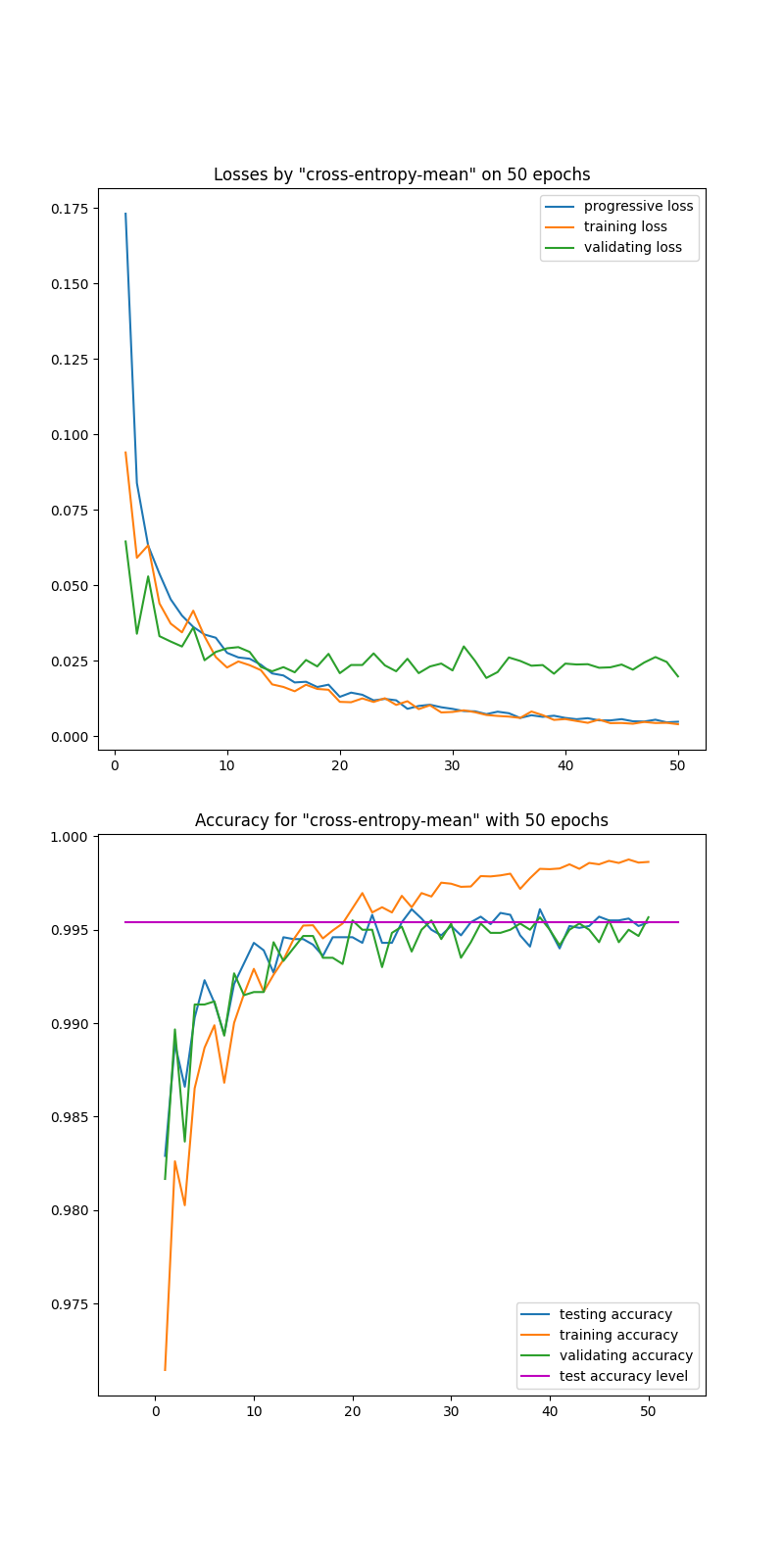} 
&
\includegraphics[height=120mm]{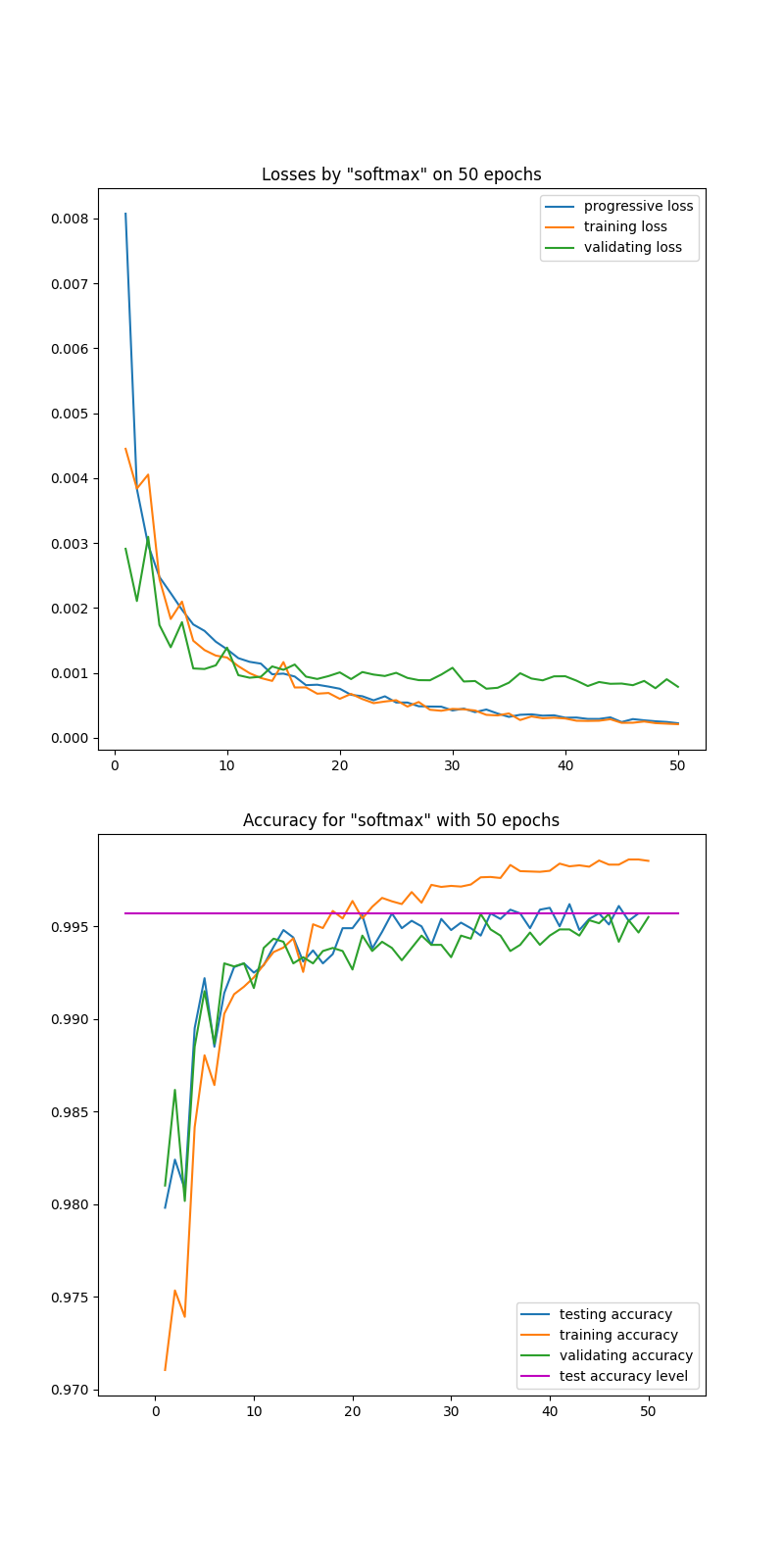}
&
\includegraphics[height=120mm]{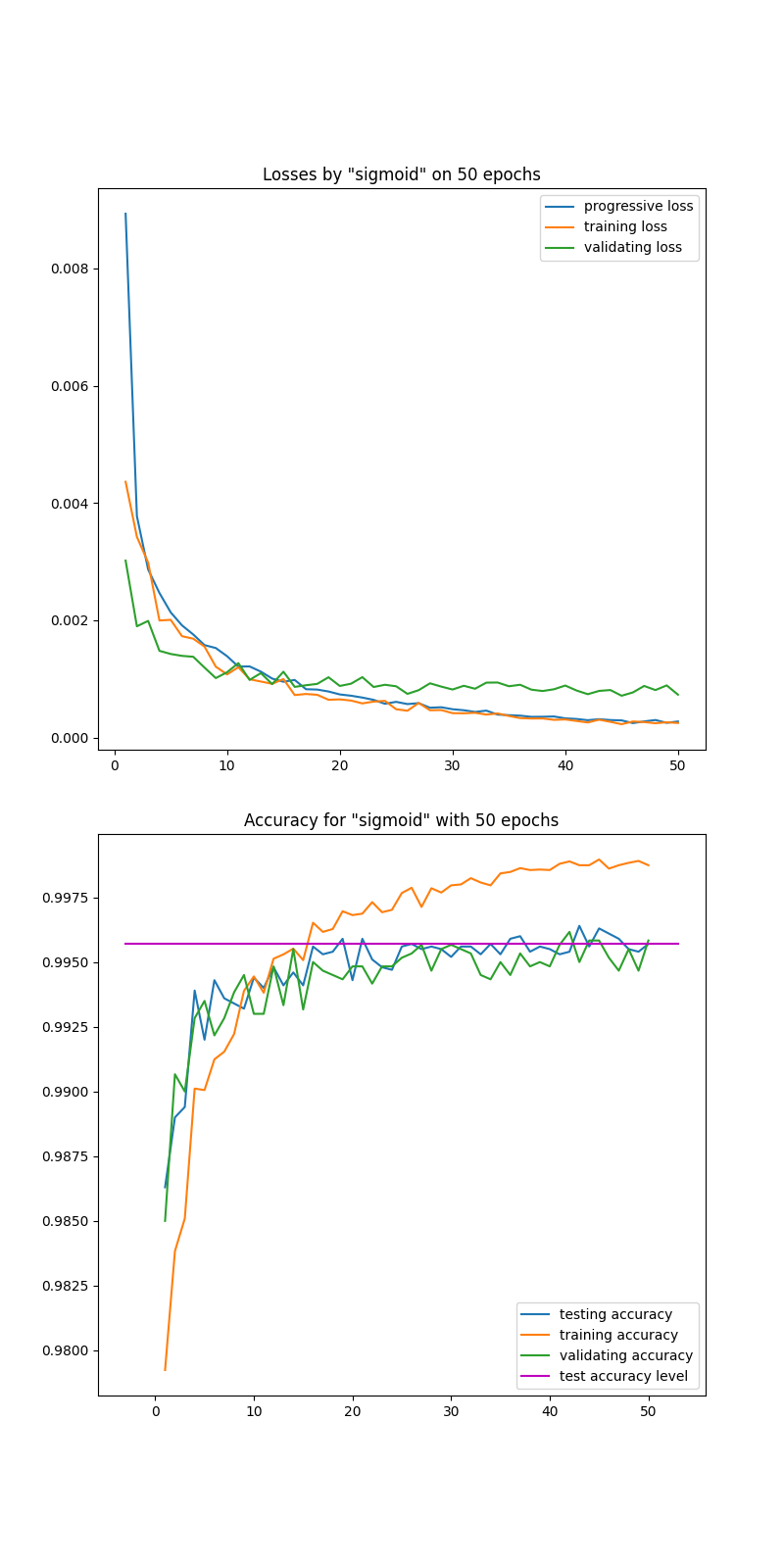} 
\end{tabular}
}
\caption{Learning curves on training and validation data for the $\cl{N}_1$ network and three models: {\tt cross-entropy-mean, softmax, sigmoid}. The horizontal reference line represents the accuracy on test data computed after the last epoch.}
\label{fig:learn-tr-val}
\end{figure}

First, let's see on three models {\tt cross-entropy-mean, softmax, sigmoid} their loss and efficiency curves obtained on training data {\tt MNIST(54K)} and on data intended solely for model validation {\tt MNIST(6K) }. These two loss curves are calculated after each epoch. We supplement them with a loss curve calculated progressively after each batch of training data (see figure \ref{fig:learn-tr-val}).

Let us note the correct course of the train loss curve with respect to progressive loss curve - both curves are close. The correct course is also for validation loss curve - the validation curve from about epoch 30 is below the training curve maintaining a significant distance. This is a sign that the model is not overfitted. This effect was achieved only after applying a moderate input image augmentation procedure.

Correct behavior of learning curves was recorded both for the modesl with entropy and for models with the ISBE unit. This also applies to classifier performance curves.

\begin{figure}[ht]
\centerline{
\begin{tabular}{ccc}
\includegraphics[height=60mm]{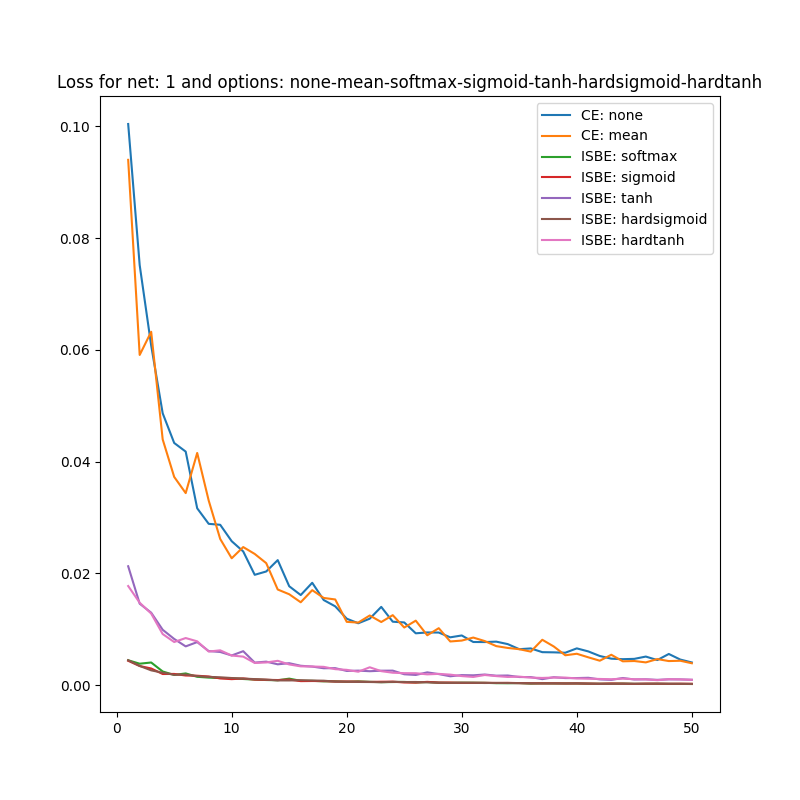} 
&
\includegraphics[height=60mm]{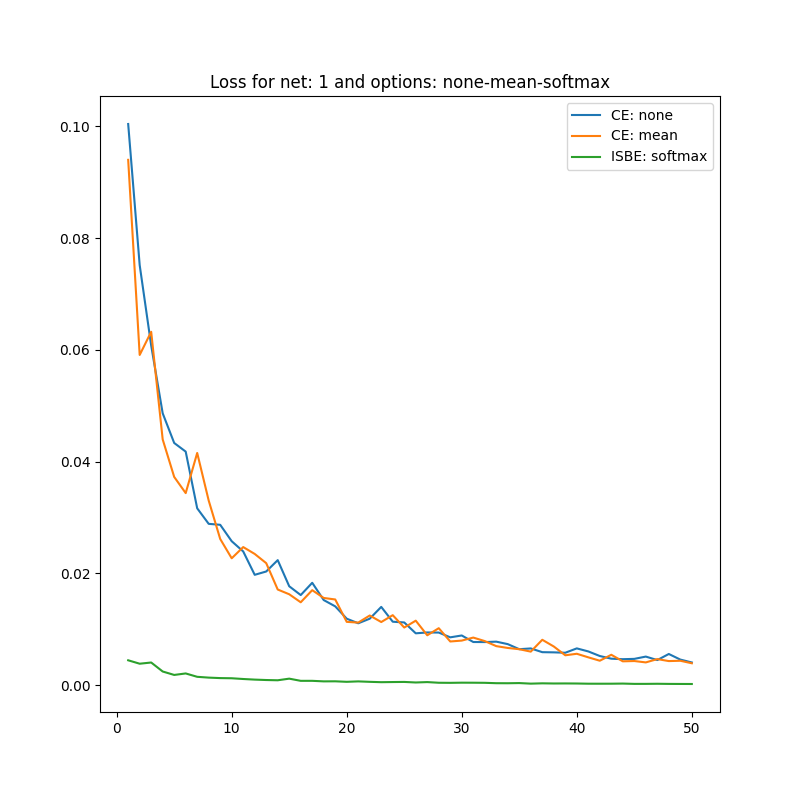}
&
\includegraphics[height=60mm]{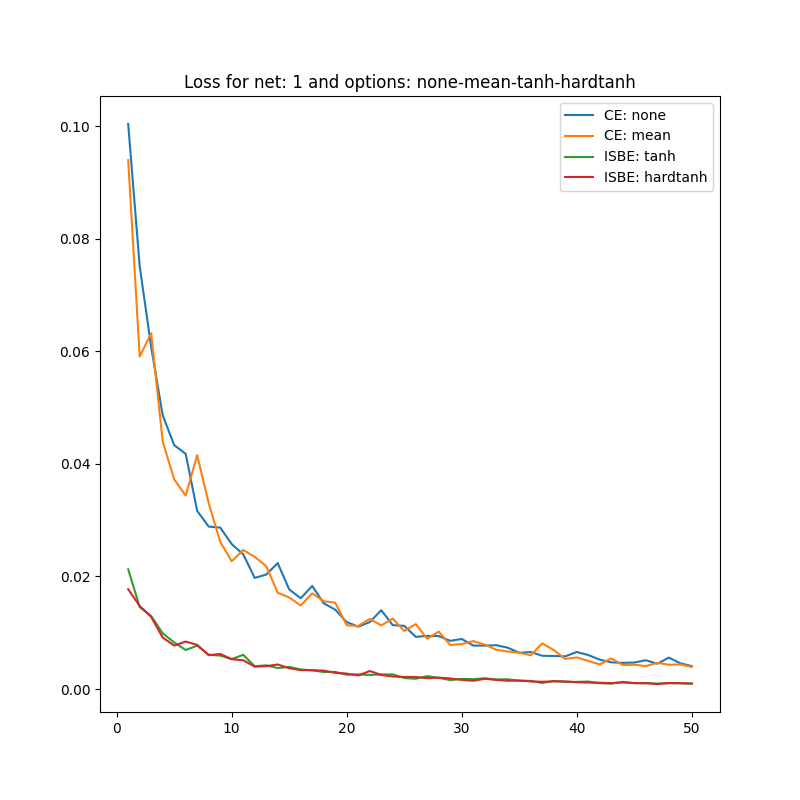} 
\\
\includegraphics[height=60mm]{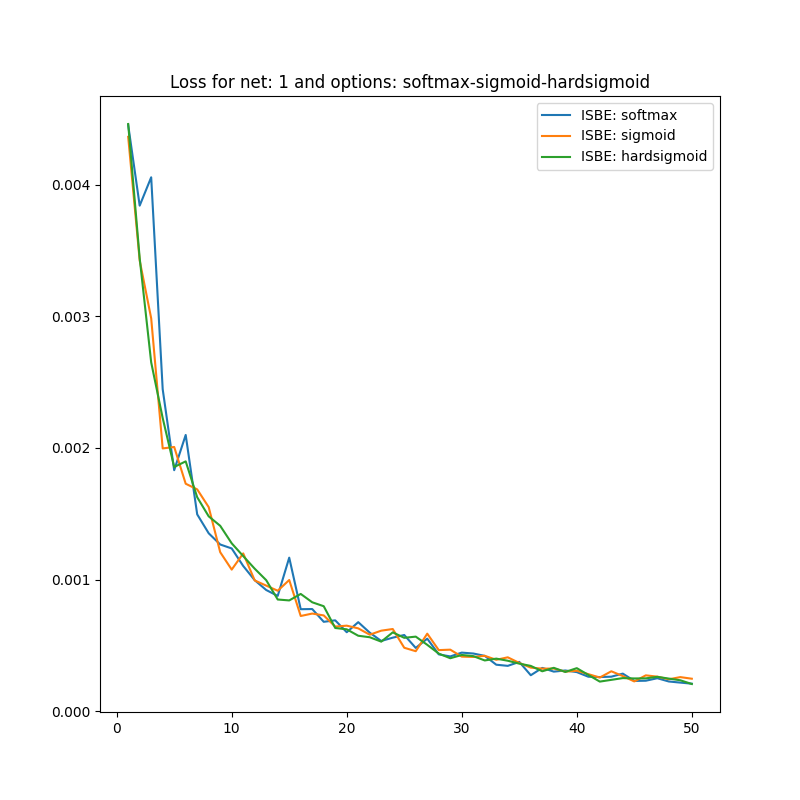} 
&
\includegraphics[height=60mm]{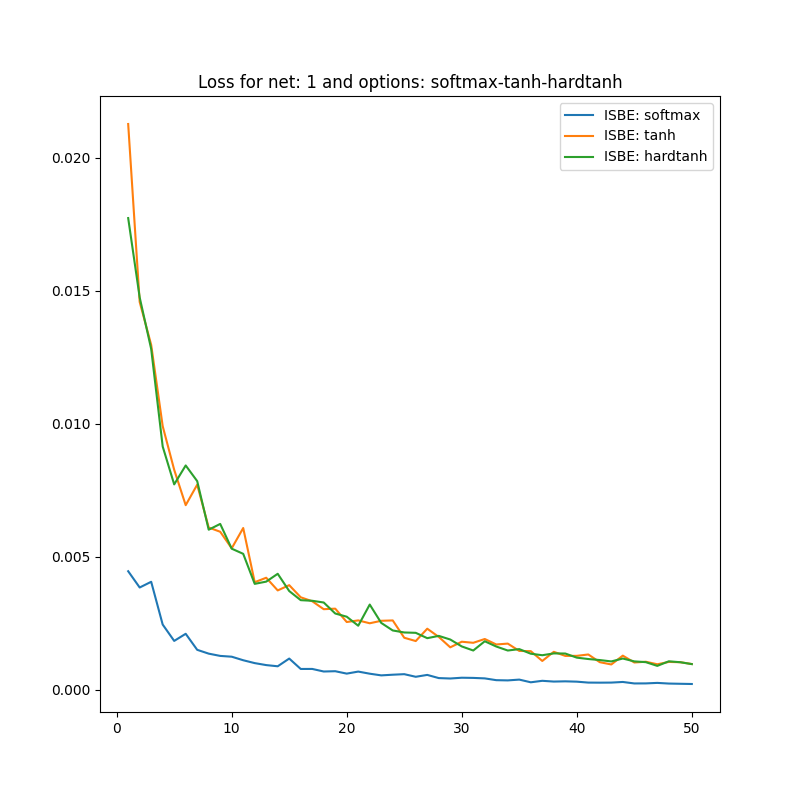}
&
\includegraphics[height=60mm]{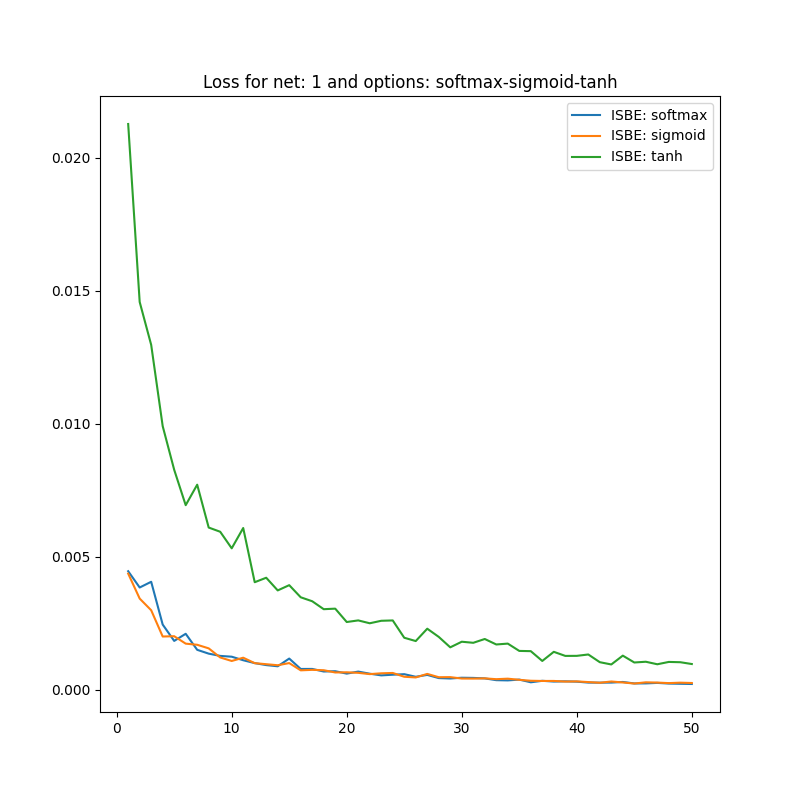} 
\end{tabular}
}
\caption{Loss charts of learning in comparisons of CE versus ISBE options. 
In the first row: (1) all options for loss functions and soft score functions; (2) {\tt CE none, CE mean} versus {\tt softmax}; (3): {\tt CE none, CE mean} versus {\tt tanh, hardtanh}. In the second row: (1) {\tt softmax} versus {\tt sigmoid, hardsigmoid}; (2) {\tt softmax} versus {\tt tanh, hardtanh}; (3) {\tt softmax} versus {\tt sigmoid, tanh}.}
\label{fig:learn}
\end{figure}

\begin{enumerate}
\item Curves of loss functions can appear together as long as the type of function is identical, which entails a similar range of variability for loss function values. One might wonder what measure of loss to adopt in the case of ISBE, since this technique, in fact, does not calculate loss values. We opt for a natural choice of mean square error for errors in soft scores:
$$
\cl{L}_{ISBE} = MSE(Y, Y^{\circ}) \eqd \frac{1}{n_b}\cdot \|Y-Y^{\circ}\|^2_2
$$
For such defined measures, it turns out that only the option of reduction by summing has a different range of variability, and therefore it is not on the figure \ref{fig:learn}.
\item In the case of classifier accuracy, a common percentage scale does not exclude placing all eight curves for each considered architecture. However, due to the low transparency of such a figure, it is also worth juxtaposing different groups of curves of the dependency $\alpha(e)$. The accuracy $\alpha$ of the classifier {\tt MNIST(60K)} is calculated on the test set {\tt MNIST(10K)}. 
\end{enumerate}

\begin{figure}[ht]
\centerline{
\begin{tabular}{ccc}
\includegraphics[height=60mm]{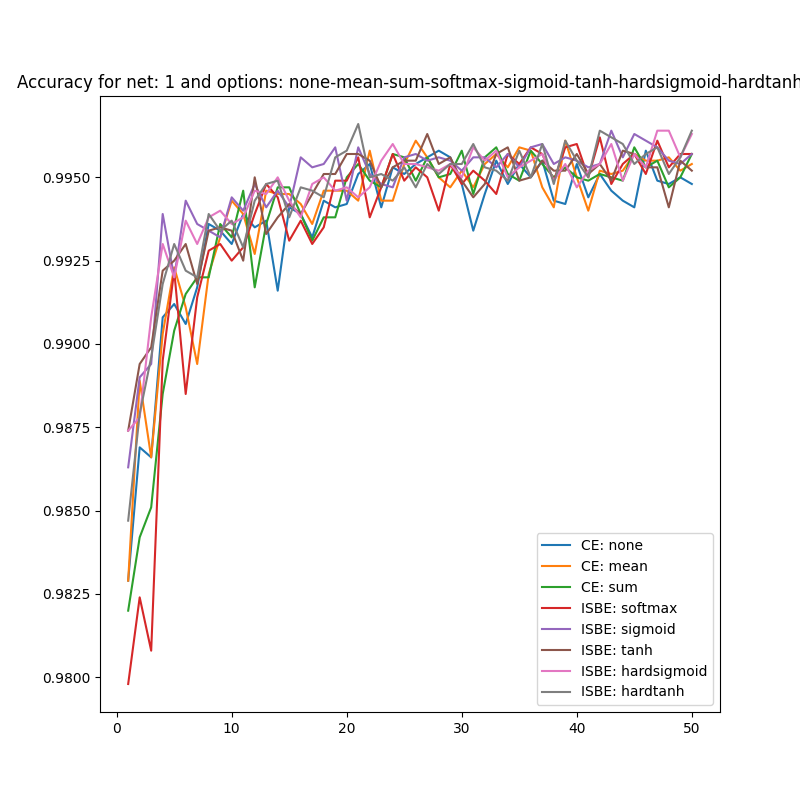} 
&
\includegraphics[height=60mm]{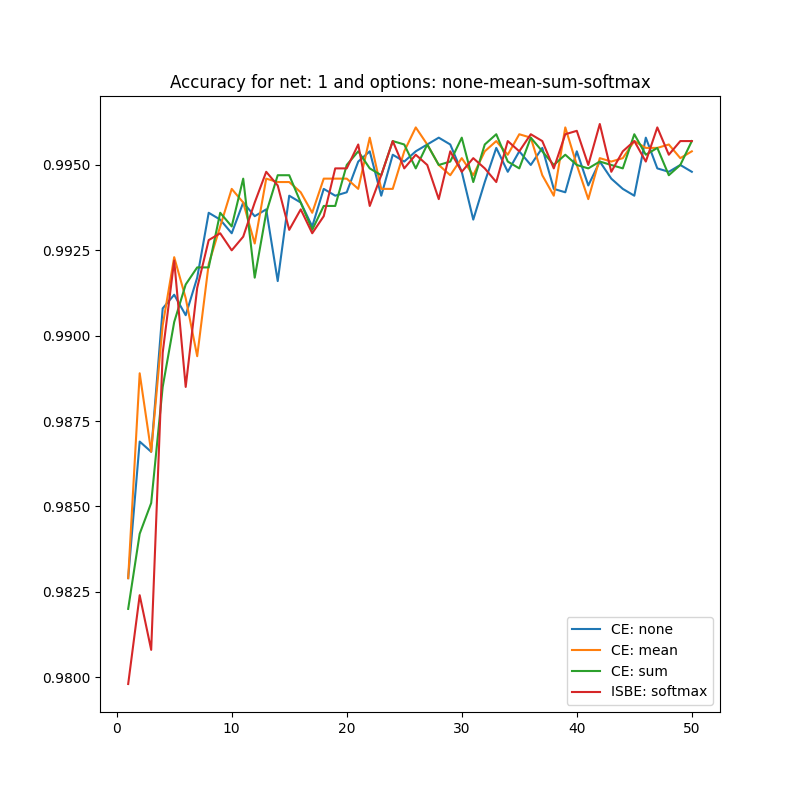}
&
\includegraphics[height=60mm]{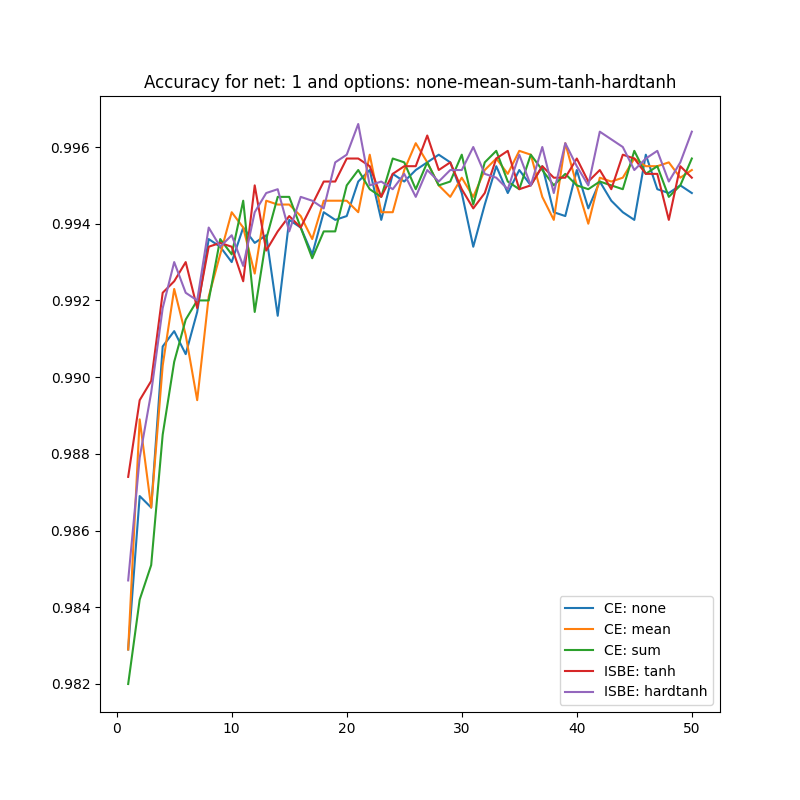} 
\\
\includegraphics[height=60mm]{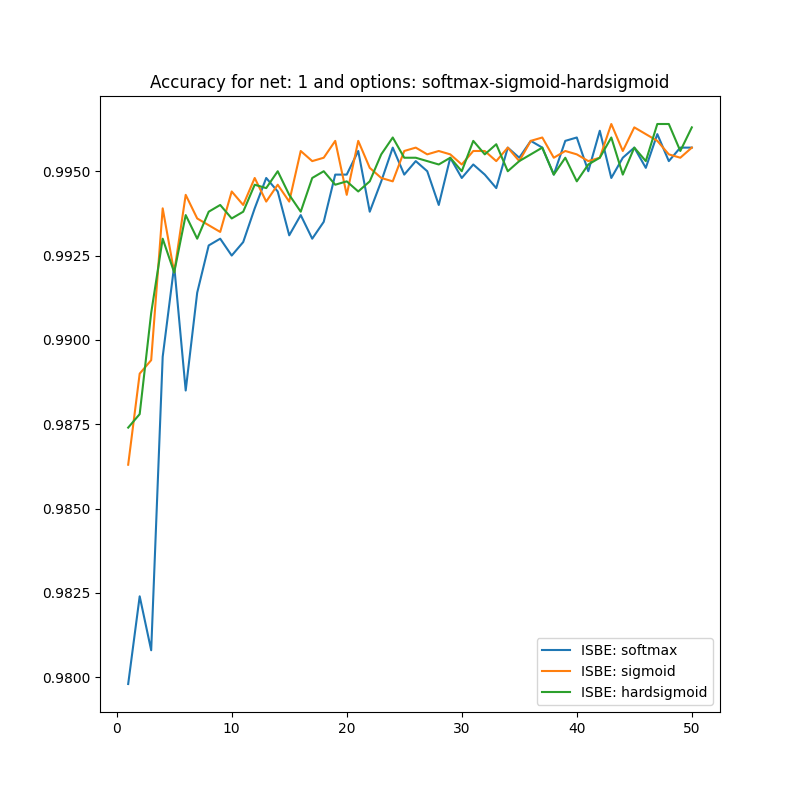} 
&
\includegraphics[height=60mm]{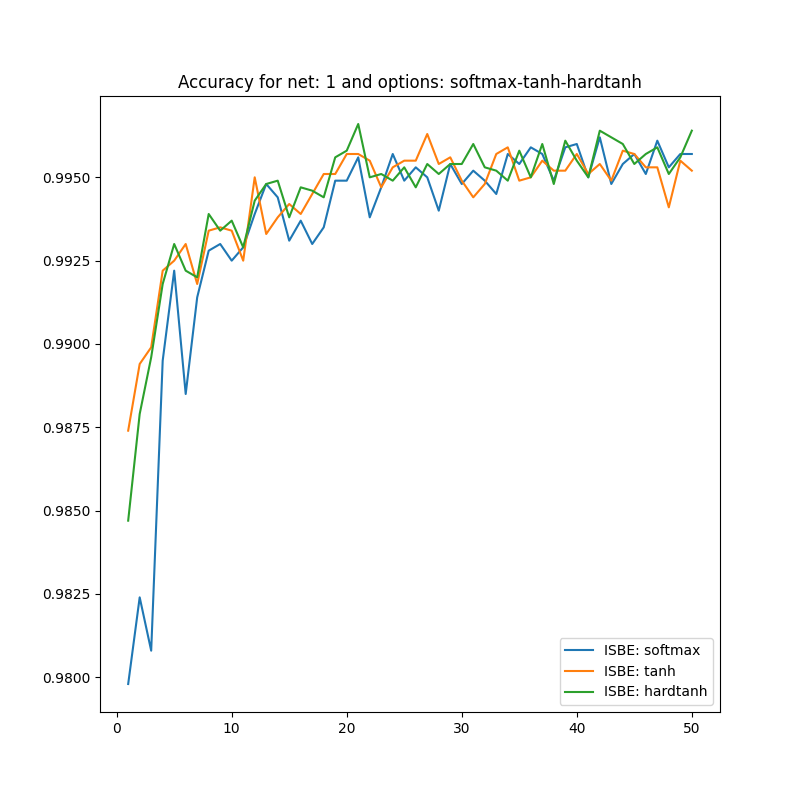}
&
\includegraphics[height=60mm]{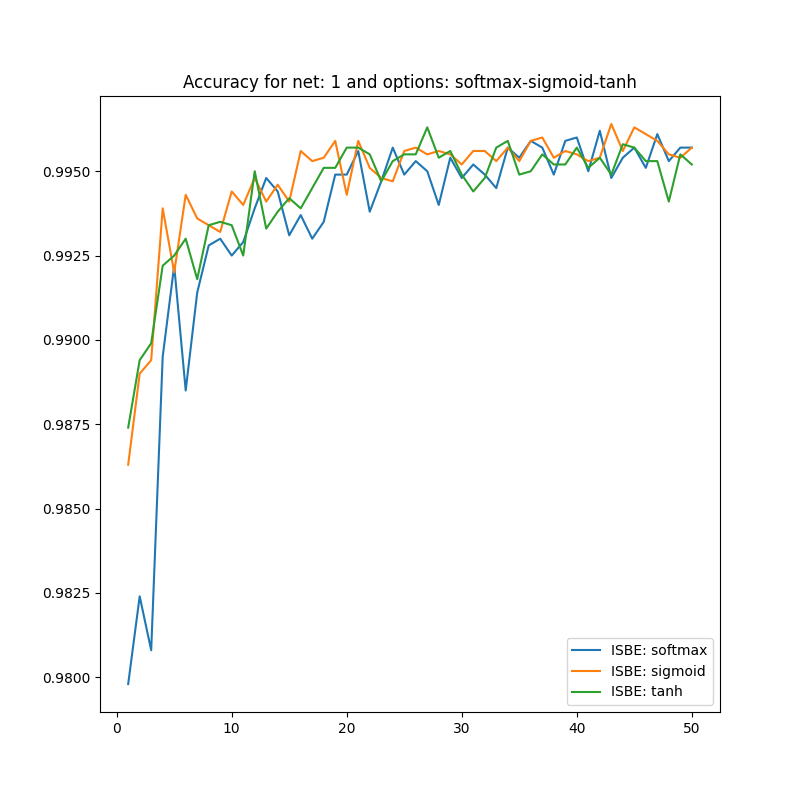} 
\end{tabular}
}
\caption{Accuracy charts of learning in comparisons of CE versus ISBE options. 
In the first row: (1) all options for loss functions and soft score functions; (2) {\tt CE none, CE mean} versus {\tt softmax}; (3): {\tt CE none, CE mean} versus {\tt tanh, hardtanh}. In the second row: (1) {\tt softmax} versus {\tt sigmoid, hardsigmoid}; (2) {\tt softmax} versus {\tt tanh, hardtanh}; (3) {\tt softmax} versus {\tt sigmoid, tanh}.}
\label{fig:accuracy}
\end{figure}

Sets of curves, which we visualize separately for architectures 
$\cl{N}_0$, $\cl{N}_1$ are:
\begin{itemize}
\item all options for loss functions (3) and soft score functions (5),
\item {\tt CE none, CE mean, CE sum} versus {\tt softmax},
\item {\tt CE none, CE mean, CE sum} versus {\tt tanh, hardtanh},
\item {\tt softmax} versus {\tt sigmoid, hardsigmoid},
\item {\tt softmax} versus {\tt tanh, hardtanh},
\item {\tt softmax} versus {\tt sigmoid, tanh}.
\end{itemize}
Due to space constraints, we show learning curves and classifier effectiveness graphs only for architecture $\cl{N}_1$ in figures \ref{fig:learn}, \ref{fig:accuracy}.

In figure \ref{fig:learn} we can clearly observe four clusters of models:
\begin{itemize}
\item {\tt CrossEntropyLoss} based  with reduction option {\tt sum} (as out of common range it was not shown),
\item {\tt CrossEntropyLoss} based with reduction options {\tt none,} and {\tt  mean},  
\item {\tt ISBE} based with normalizations  to range $[0,1]$ including functions\\
 $SoftMax,$ $Sigmoid,$ and $HardSigmoid$,
\item {\tt ISBE} based with normalizations  to range $[-1,1]$ including functions $Tanh,$ and $HardTanh$.
 \end{itemize}
 Within a cluster, the loss curves behave very similarly.
 Interestingly, the loss curves in ISBE-based clusters tend to the same value greater than zero. In contrast, cross-entropy-based curves also tend to the same limit. However it is clearly greater than ISBE one.
 
Now, we will pay more attention to test learning curves. We generate test learning curves on the full set of test data {\tt MNIST(10K)}. After each epoch, one point is scored towards the test learning curve. We will show these curves in several comparative contexts.

In  the case of classifier accuracy curves (see figure \ref{fig:accuracy}), the variances in the clusters described above are smaller than in the union of clusters. Close to the final epochs, all curves tend to be chaotic within the range of $(99.4,99.7)$. 

\begin{figure}[ht]
\centerline{
\begin{tabular}{ccc}
\includegraphics[height=60mm]{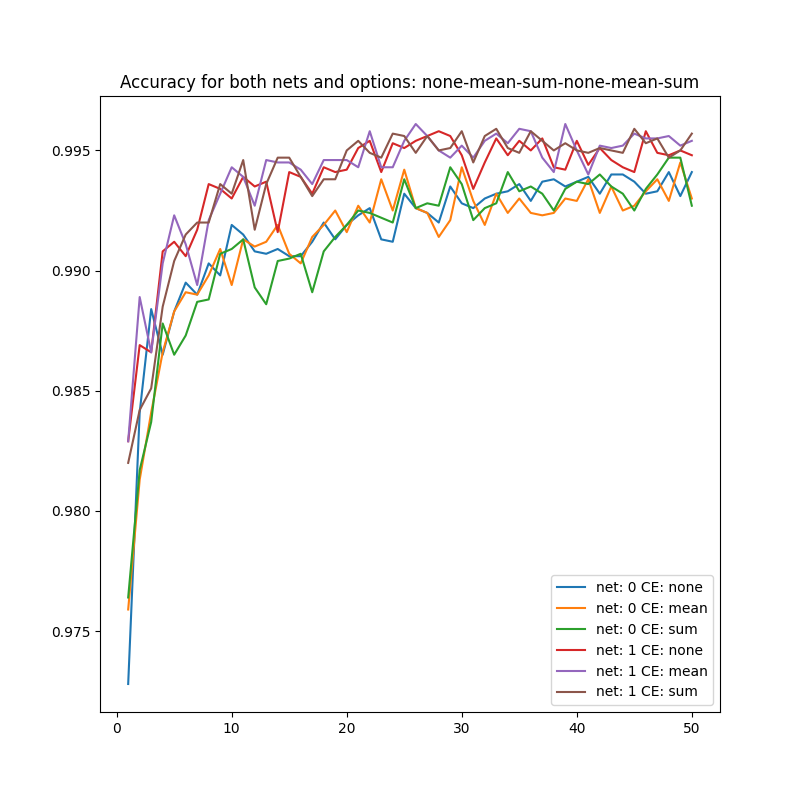} 
&
\includegraphics[height=60mm]{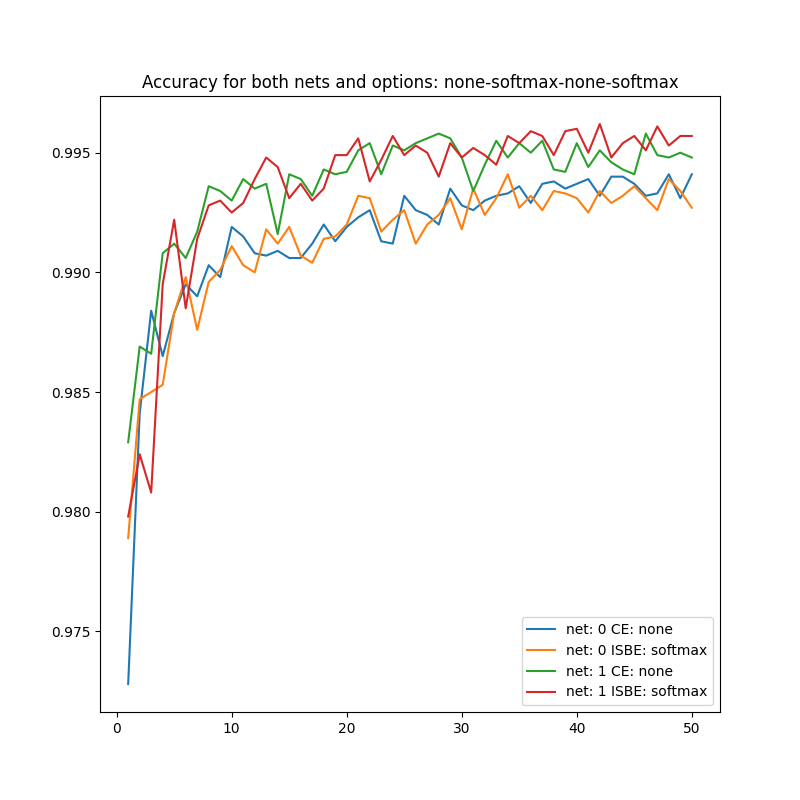}
&
\includegraphics[height=60mm]{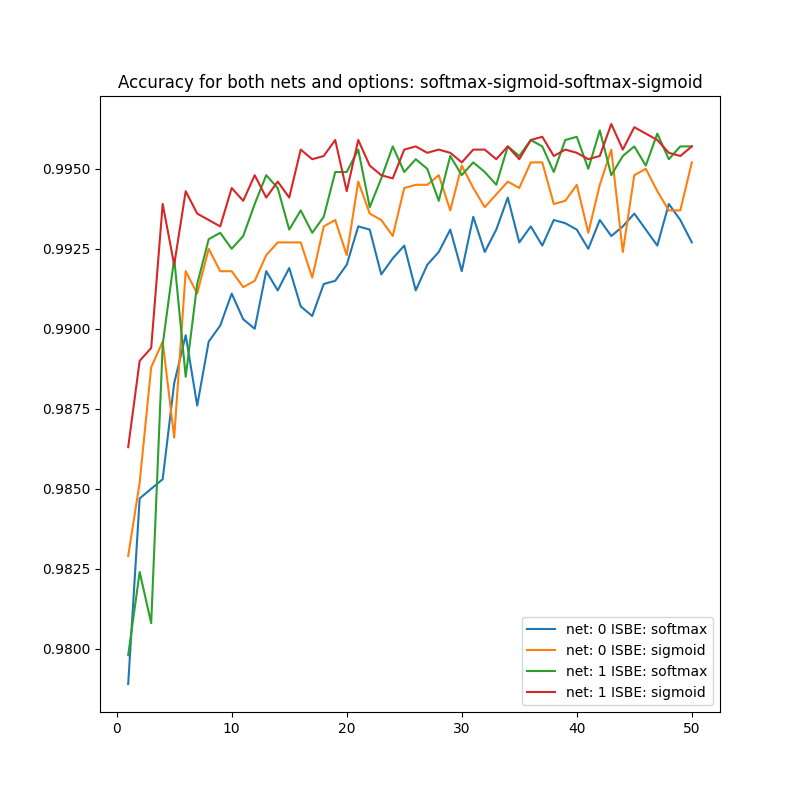} 
\\
\includegraphics[height=60mm]{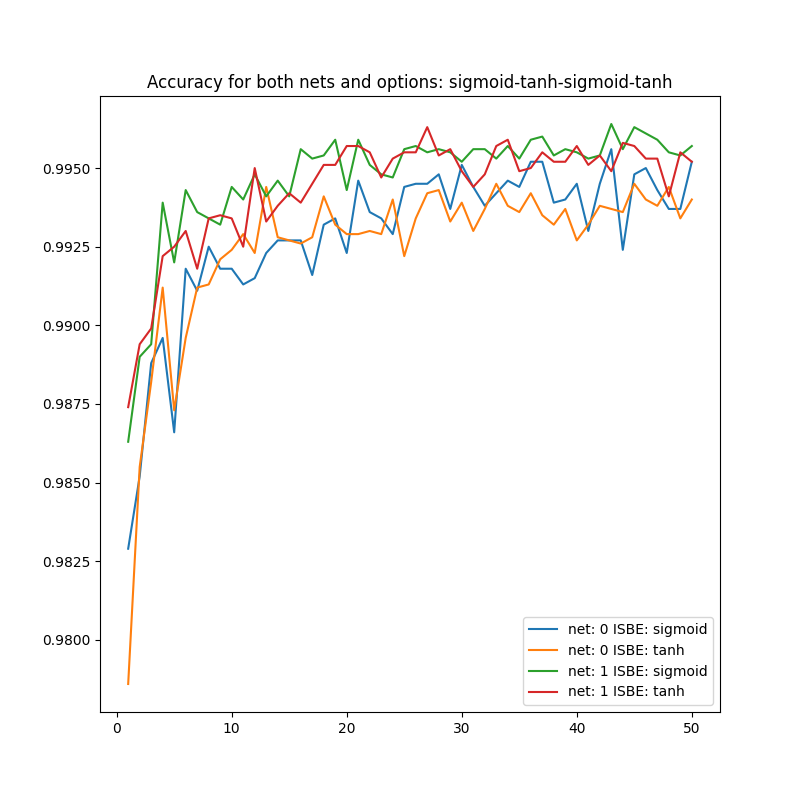} 
&
\includegraphics[height=60mm]{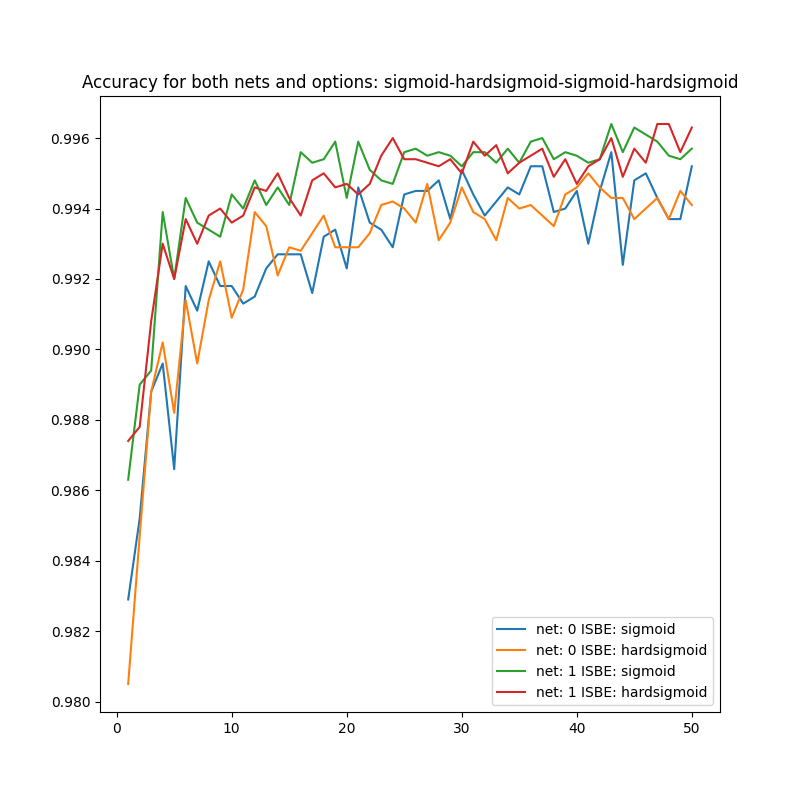}
&
\includegraphics[height=60mm]{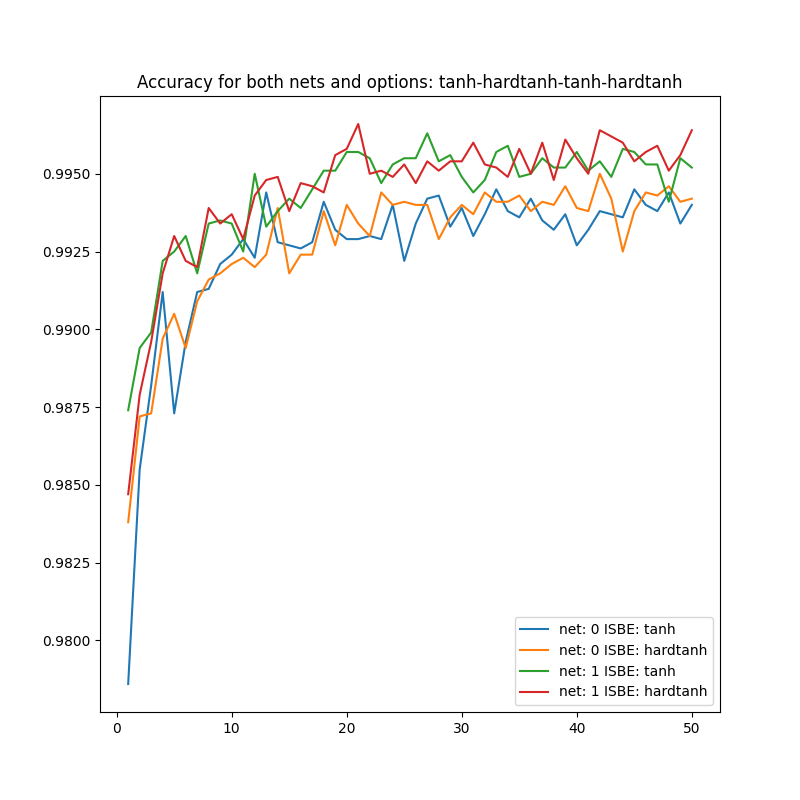} 
\end{tabular}
}
\caption{Accuracy charts of learning in comparisons of CE versus ISBE options and architecture $\cl{N}_0$ versus $\cl{N}_1$. 
In the first row: (1) {\tt CE none, CE mean, CE sum}; (2) {\tt CE none, softmax}; (3): {\tt softmax, sigmoid}. In the second row: (1) {\tt sigmoid, tanh}; (2) {\tt sigmoid, hardsigmoid} ; (3) {\tt tanh, hardtanh}.}
\label{fig:accuracy-both}
\end{figure}

Visualizing the effectiveness of classifiers for different architectures of different complexities, although more obvious, also has research value (see figure \ref{fig:accuracy-both}):
\begin{itemize}
\item {\tt CE none, CE mean, CE sum} from $\cl{N}_0$ versus 
      {\tt CE none, CE mean, CE sum} from $\cl{N}_1$,
\item {\tt CE none, softmax} from $\cl{N}_0$ versus
      {\tt CE none, softmax} from $\cl{N}_1$,
\item {\tt softmax, sigmoid} from $\cl{N}_0$ versus
      {\tt softmax, sigmoid} from $\cl{N}_1$,
\item {\tt sigmoid, tanh} from $\cl{N}_0$ versus
      {\tt sigmoid, tanh} from $\cl{N}_1$,
\item {\tt sigmoid, hardsigmoid} from $\cl{N}_0$ versus
      {\tt sigmoid, hardsigmoid} from $\cl{N}_1$,
\item {\tt tanh, hardtanh} from $\cl{N}_0$ versus
      {\tt tanh, hardtanh} from $\cl{N}_1$.
\end{itemize}

The figure \ref{fig:accuracy-both} shows the better performance of $\cl{N}_1$ than $\cl{N}_0$. Moreover, we can observe slightly more stable behaviour for ISBN-based curves than for cross-entropy-based.

\section{Conclusions}

Cross-entropy CE as a loss function {\it owes much to normalization} performed by the {\tt SoftMax} activation function. In the backward gradient backpropagation phase, only this activation, through perfect linearization, can prevent the explosion or suppression of the gradient originating from CE. What we call the {\it softmax trick}, as a mathematical phenomenon is explained by the theory presented in Appendix A. There is a proof that such linearization can only be realized by a function $F:\bb{R}^K\ra\bb{R}^K$ with a Jacobian identical to that of the {\tt SoftMax} function. In turn, such a Jacobian can only be  by {\it relocated} versions of the {\tt SoftMax} function. 

For another research there are left practical aspects of more general theorem \ref{th:strick2} implying that dilated and relocated versions of SoftMax, are the only ones having the property of {\it dilated softmax trick}.

Should we, therefore, celebrate this unique relationship between activation and cost function? In this work, I have shown that it is rather beneficial to use the final effect of the action of this pair, namely the linear value equal to $Y-Y^{\circ}$, which can be calculated without their participation. This is exactly what the {\tt ISBE} unit does - it calculates the soft score vector in the forward step to return in backward step its error from the target score.

To determine the normalized score, the {\tt ISBE} unit can use not only the {\tt SoftMax} function, as it is not necessary to meet the unity condition, i.e., to ensure a probability distribution as scores of the trained classifier. At least four other activation functions {\tt Sigmoid, Tanh} and their hard versions {\tt HardSigmoid} and {\tt HardTanh} perform no worse. The choice of these final activations was rather a matter of chance, so researchers face further questions. How to normalize raw scores and how to appropriately represent (encode) class labels in relation to this normalization, so as not to degrade the classifier's results? What properties should such normalization functions have? Experiments suggest that meeting the Lipschitz condition in the vicinity of zero may be one of these properties.

The theoretical considerations presented prove that the {\tt ISBE} unit in the process of deep model learning correctly simulates the behavior of the {\tt CrossEntropy} unit preceded by the {\tt SoftMax} normalization.

The experiments showed that the {\tt ISBE} unit saves the time time of  forward and backward stage up to $3\%$, and the effectiveness of the classifier model remains unchanged within the margin of statistical error. 

On the other hand, the examples of code fragments showed that the programmer's time {\it spent} on introducing the {\it ISBE} technique to his/her program instead of {\tt CrossEntropyLoss} is negligible.

\appendix

\section{Cross-Entropy and Softmax Trick Properties}

The discrete cross-entropy function $CE$ of a target discrete probability distribution $y^{\circ}\in[0,1]^K,\ \sum_iy^{\circ}_i=1$, relative to the distribution calculated by the classifier $y\in(0,1)^K,\ \sum_iy_i=1$, is defined by the formula $CE(y^{\circ},y) \eqd -\sum_iy^{\circ}_i\ln y_i$. 

The gradient of $CE$ with respect to $y$ is $\od{CE}{y}(y^{\circ},y)=-y^{\circ}\div y,$ where the operation $w = u\div v$ denotes division by components of vectors $u,v$, i.e., $w_i \eqd u_i/v_i$, $i\in[K]$. The notation $[K]$ refers to a sequence of $K$ indices, such as $(1,\dots,K)$ or $(0,1,\dots,K-1)$.

In the special case when the vector $y\in(0,1)^K$ is calculated based on the vector $x\in\bb{R}^K$ according to the formula $y_i = SoftMax(x)_i=e^{x_i}/\sum_ke^{x_k}$, the gradient of $CE$ with respect to $x$ has a particularly simple resultant formula. Its simplicity was the reason for the term {\it softmax trick}:
\begin{equation*}\label{eq: soft-trick}
y=SoftMax(x) \lra \od{CE}{x}(y^{\circ},y(x)) \xeq{\text{\it softmax trick}} y-y^{\circ}
\end{equation*}

Some authors (\cite{blog on strick}) also use the term {\it softmax trick} for that part of the proof showing that the derivative of the natural logarithm of the sum of functions $e^{x_i}$ equals to the $SoftMax$ function: 
$$
\begin{array}{l}
\ds y_i\eqd \frac{e^{x_i}}{\sum_ke^{x_k}} \lra
\ds\sp{\left[\ln(\sum_ke^{x_k})\right]}{x_i}=y_i \lra \\[10pt]
\begin{array}{rcl}
\ds\sp{CE(y^{\circ},y(x))}{x_i} &=& \ds
\sp{\left[\sum_jy^{\circ}_j\ln(\sum_ke^{x_k})-\sum_jy^{\circ}_j\ln e^{x_j}\right]}{x_i}\\
&=& y_i\overbrace{\sum_jy^{\circ}_j}^{=1}-y^{\circ}_i = y_i-y^{\circ}_i\quad\Box
\end{array}
\end{array}
$$
In matrix notation (\cite{matrix compute}), the property of {\it softmax trick} has a longer proof, as we first need to calculate the Jacobian of the $SoftMax$ function (\cite{jacobian}), which is $\sp{y}{x} = \diag{y}-y\tp{y}$. Then
\begin{equation*}\label{eq:jacob2strick}
 \begin{array}{rcl}
 \ds\sp{[CE(y^{\circ},y(x))]}{x} &=& \ds\sp{y}{x}\cdot \od{CE}{y}(y^{\circ},y) =\left(\diag{y}-y\tp{y}\right)\left(-y^{\circ}\div y\right)\\[8pt] 
 &=& \ds
 y\underbrace{\tp{(y\div y)}y^{\circ}}_{\tp{\mathbf{1}}_Ky^{\circ}=1} - \underbrace{\diag{y \div  y}}_{I_K} y^{\circ}
 = y-y^{\circ}
 \end{array}  
\end{equation*}

An interesting question arises: Is it only the $SoftMax$ function that has such a property? It seems possible, as for any differentiable function $F:\bb{R}^n\ra\bb{R}^n$ the starting point is similar:
\begin{equation*}\label{eq:soft-trick-F}
\sp{\left[CE(y^{\circ},\underbrace{F(x)}_{y})\right]}{x}
= \sp{F(x)}{x}\overbrace{\sp{[CE(y^{\circ},y)]}{y}}^{-y^{\circ}\div y} = \sp{F(x)}{x}(-y^{\circ}\div y)
\end{equation*}
The following theorem fully characterizes functions that have the {\it softmax trick} property.
 
\begin{theorem}[on the properties of softmax trick]\label{th:strick}$ $

For a differentiable function $F:\bb{R}^K\ra\bb{R}^K$, the following three properties are equivalent:
\begin{enumerate}
\item $F$ is a {\it SoftMax function with relocation}, if there exists a reference point $c\in\bb{R}^K$, such that for every $x\in\bb{R}^K$:
\begin{equation*}\label{eq:smax-reloc}
y = F(x) = SoftMax(x-c)\,,
\end{equation*}
\item $F$ has a {\it softmax-type Jacobian}, if for every $x\in\bb{R}^K$:
\begin{equation}\label{eq:jacobian}
Jacobian(F)(x) \eqd \sp{F(x)}{x} = \diag{y}-y\tp{y}\,, \text{ where } y=F(x)\,,
\end{equation}
\item $F$ possesses the {\it softmax trick} property, if
for every $y^{\circ}\in[0,1]^K, x\in(0,1)^K$:
\begin{equation} \label{eq:strick}
\sp{F(x)}{x}(-y^{\circ}\div y) = y-y^{\circ}\,, \text{ where } y=F(x)\,.
\end{equation}
\end{enumerate}
\end{theorem}

\begin{proof}[Proof of Theorem \ref{th:strick}]$ $

We prove the implications in the following order: 
$(2)\lra(3),\ (3)\lra(2)$,\  $(1)\lra(2),\ (2)\lra(1)\,.$
\begin{itemize}
\item Proof of implication $(2)\lra(3)$: 

If the Jacobian $\sp{F(x)}{x}$ of the function $F$ is of the {\it softmax} type, then for $y=F(x)$:
$$\sp{y}{x}=\left(\diag{y}-y\tp{y}\right)\left(-y^{\circ}\div y\right)
=
 y\underbrace{\tp{(y\div y)}y^{\circ}}_{\tp{\mathbf{1}}_Ky^{\circ}=1} - \underbrace{\diag{y \div  y}}_{I_K} y^{\circ}
 = y-y^{\circ}\quad\Box
$$
\item Proof of implication $(3)\lra(2)$:
Denote the axis unit vector $j$ by $e_j\in\bb{R}^K$. 
Then $(e_j)_i=\delta_{ij}$. Substitute into property \eqref{eq:strick} the target score vector $y^{\circ}\eqd e_j$. Then 
$$
y_i-(e_j)_i=\tp{\left(\sp{F(x)}{x_i}\right)}(-e_j\div y)= \sum_{k\in[K]}\sp{y_k}{x_i}\cdot \left(\frac{-\delta_{kj}}{y_k}\right) = 
\sp{y_j}{x_i}\cdot \left(\frac{-1}{y_j}\right)
$$
Therefore $\sp{y_j}{x_i} = ((e_j)_i-y_i)y_j \xeq{(e_j)_i=\delta_{ij}} (\delta_{ij}-y_i)y_j$.
Thus, $\sp{y}{x} = \diag{y}-y\tp{y}\,.$\quad$\Box$
\item Proof of implication $(1)\lra(2)$:

If $\ds y_j\eqd\frac{e^{x_j-c_j}}{\sum_{k\in[K]}e^{x_k-c_k}}$, then\\[5pt]
$
\ds\sp{y_{j}}{x_{i}} = \left\{
\begin{array}{l}
\ds\frac{-e^{x_{j}-c_j}\cdot e^{x_{i}-c_i}}
{\left(
\sum_{k\in[K]}e^{x_{k}-c_k}
\right)^2}
= -y_{i}\cdot y_{j},\ \text{when } i\neq j\\[20pt]
\ds\frac{e^{x_{i}-c_i}\cdot \left(\sum_{k\in[K]}
e^{x_{k}-c_k}\right)-e^{x_{i}-c_i}\cdot e^{x_{i}-c_i}}
{\left(\sum_{k\in[K]}e^{x_{k}-c_k}\right)^2}\\[20pt]
= y_{i}-y_{i}^2 = (1-y_{i})\cdot y_{i},\ \text{when } i=j
\end{array}
\right.
$

The general formula is $\ds\sp{y_{j}}{x_{i}}=(\delta_{ij}-y_i)y_j\,.$ Therefore:
$$
\left(\sp{y}{x}\right)_{ij} \eqd \sp{y_{j}}{x_{i}}= \delta_{ij}y_j-y_iy_j =
\left(\diag{y}\right)_{ij} - \left(y\tp{y}\right)_{ij}\,.
$$
Hence, $\sp{y}{x}=\diag{y}-y\tp{y}\quad\Box$

\item Proof of implication $(2)\lra(1)$:\\
If $\sp{y_{i}}{x_{i}}=(1-y_{i})\cdot y_{i}$, then the diagonal of the Jacobian matrix gives us differential equations (\cite{diff-equations}), from which we can determine the general form of the function $y_i(x)$, $i\in[K]:$
$$
\begin{array}{l}
\ds\frac{\partial y_i}{(1-y_{i})\cdot y_{i}} =\partial x_i \lra
\int\left(
\ds\frac{1}{y_i}+\frac{1}{1-y_i}\right)\partial y_i = \int\partial x_i\\[15pt] 
\ds\lra\ln\frac{y_i}{1-y_i} = x_i + C(k\neq i) \lra\ds
y_i = \ds\frac{e^{x_i}}{\underbrace{e^{x_i}+e^{-C(k\neq i)}}_{z_i}}\\[10pt]
\lra \ds y_i = \frac{e^{x_i}}{z_i}, \text{ where } z_i=z_i(x_1,\dots,x_K)>0\\[10pt]
\lra \ln{z_i} = x_i-\ln y_i
\end{array}
$$ 
Now we calculate the partial derivatives $\sp{\ln z_j}{x_i}$. 
If $i\neq j$, then
$$
\sp{[\ln z_j]}{x_i} = \sp{[x_j-\ln y_j]}{x_i} = -\frac{1}{y_j}\overbrace{\sp{y_j}{x_i}}^{-y_jy_i} = y_i 
$$
For $i=j$, the result is the same:
$$
\sp{[\ln z_i]}{x_i} = \sp{[x_i-\ln y_i]}{x_i} = 1 -\frac{1}{y_i}\overbrace{\sp{y_i}{x_i}}^{(1-y_i)y_i} = 1-\frac{(1-y_i)y_i}{y_i} = y_i 
$$
Therefore, for a fixed $i$, e.g., for $j=1$, we have $K$ equalities: 
$\sp{[\ln z_1]}{x_i}=\sp{[\ln z_j]}{x_i}=y_i$ for any $j,i\in[K]$.
This means that vector fields for each pair of functions $\ln z_1$ and $\ln z_j$ are identical.

Integrating these fields yields the same functions up to a constant $c_j$:
$\ln z_j = \ln z_1 + c_j,$ $j\in[K]$.  
Consequently, $z_j=z_1\cdot e^{c_j}$, $j\in[K]$, and therefore 
$y_j = \ds\frac{e^{x_j}}{z_1e^{c_j}} = \frac{e^{x_j-c_j}}{z_1}$. 
From the {\it unity} condition, we can now determine the value of $z_1$:
$$
1=\sum_{k\in[K]}y_k=\sum_{k\in[K]}
\frac{e^{x_k-c_k}}{z_1} 
\lra z_1 = \sum_{k\in[K]}e^{x_k-c_k} \lra y_j =  \ds\frac{e^{x_j-c_j}}{\ds\sum_{k\in[K]}e^{x_k-c_k}}\,.
$$
\end{itemize}
\end{proof}

The theorem \ref{th:strick} can be generalized to  dilated version of the {\it SoftMax} function. It is reformulated in the form of theorem \ref{th:strick2}. The proof of the generalized theorem can be easily generalized, too.

\begin{theorem}[on the properties of dilated softmax trick]\label{th:strick2}$ $

For a differentiable function $F:\bb{R}^K\ra\bb{R}^K$, the following three properties are equivalent:
\begin{enumerate}
\item $F$ is a {\it SoftMax function with dilation and relocation}, if there exists a reference point $c\in\bb{R}^K$, and dilation vector $d\in\bb{R}^K$, $\forall i, d_i\neq 0$ such that for every $x\in\bb{R}^K$:
\begin{equation*}\label{eq:smax-reloc2}
y = F(x) = SoftMax(d\odot x-c)\,,
\end{equation*}
\item $F$ has a {\it dilated softmax-type Jacobian}, if there exists a dilation vector $d\inv{K}$ such that for every $x\in\bb{R}^K$:
\begin{equation*}\label{eq:jacobian2}
Jacobian(F)(x) \eqd \sp{F(x)}{x} = d\odot\left(\diag{y}-y\tp{y}\right)\,, \text{ where } y=F(x)\,,
\end{equation*}
\item $F$ possesses the {\it dilated softmax trick} property, if there exists a dilation vector $d\inv{K}$ such that for every $y^{\circ}\in[0,1]^K, x\in(0,1)^K$:
\begin{equation*} \label{eq:strick2}
\sp{F(x)}{x}(-y^{\circ}\div y) = d\odot(y-y^{\circ})\,, \text{ where } y=F(x)\,.
\end{equation*}
\end{enumerate}
\end{theorem}

\end{document}